\documentclass{article}
\usepackage{a4wide}

\usepackage{graphicx}

\usepackage{authblk}

\usepackage{hyperref}
\usepackage{url}
\usepackage{listings}

\usepackage{mathbbol}

\usepackage{listings}

\usepackage[utf8]{inputenc}
\usepackage{caption}
\usepackage{subcaption}
\usepackage{booktabs}

\usepackage{algorithm,algorithmicx,algpseudocode}

\usepackage{amsmath,amsfonts,bm}

\usepackage{amsthm}
\newtheorem{theorem}{Theorem}[section]
\newtheorem{corollary}[theorem]{Corollary}

\newtheorem{remark}[theorem]{Remark}

\def\eqref#1{equation~\ref{#1}}

\def\1{\bm{1}}

\DeclareMathAlphabet{\mathsfit}{\encodingdefault}{\sfdefault}{m}{sl}
\SetMathAlphabet{\mathsfit}{bold}{\encodingdefault}{\sfdefault}{bx}{n}

\title{Neural Integral Equations}

\author[1]{Emanuele Zappala}
\author[2]{Antonio Henrique de Oliveira Fonseca}
\author[3]{Josue Ortega Caro}
\author[4]{Andrew Henry Moberly}
\author[5]{Michael James Higley}
\author[6]{Jessica Cardin}
\author[7]{David van Dijk}

\affil[1]{Idaho State University and Yale University, emanuelezappala@isu.edu}
\affil[2]{Yale University, antonio.fonseca@yale.edu}
\affil[3]{Yale University, josue.ortegacaro@yale.edu}
\affil[4]{Yale University, andrew.moberly@yale.edu}
\affil[5]{Yale University, m.higley@yale.edu}
\affil[6]{Yale university, jess.cardin@yale.edu}
\affil[7]{Yale University, david.vandijk@yale.edu}

\date{}

\begin{document}

\maketitle

\begin{abstract}
    Nonlinear operators with long-distance spatiotemporal dependencies are fundamental in modeling complex systems across sciences, yet learning these nonlocal operators remains challenging in machine learning. Integral equations (IEs), which model such nonlocal systems, have wide-ranging applications in physics, chemistry, biology, and engineering.
    We introduce Neural Integral Equations (NIE), a method for learning unknown integral operators from data using an IE solver. To improve scalability and model capacity, we also present Attentional Neural Integral Equations (ANIE), which replaces the integral with self-attention.
    Both models are grounded in the theory of second-kind integral equations, where the indeterminate appears both inside and outside the integral operator.
    We provide theoretical analysis showing how self-attention can approximate integral operators under mild regularity assumptions, further deepening previously reported connections between transformers and integration, and deriving corresponding approximation results for integral operators. Through numerical benchmarks on synthetic and real-world data — including Lotka-Volterra, Navier-Stokes, and Burgers' equations, as well as brain dynamics and integral equations — we showcase the models' capabilities and their ability to derive interpretable dynamics embeddings. Our experiments demonstrate that ANIE outperforms existing methods, especially for longer time intervals and higher-dimensional problems.
    Our work addresses a critical gap in machine learning for nonlocal operators and offers a powerful tool for studying unknown complex systems with long-range dependencies.
\end{abstract}

\section{Introduction}\label{sec:Intro}

Integral equations (IEs) are functional equations where the indeterminate function appears under the sign of integration \cite{stech1981integral}. The theory of IEs has a long history in pure and applied mathematics, dating back to the 1800's, and it is thought to have started with Fourier's Theorem \cite{groetsch2007integral}. One other early application of IEs, was found in the Dirichelet's problem (a PDE), which was originally solved through its integral formulation. Subsequent studies, carried out by Fredholm, Volterra, and Hilbert and Schmidt, have significantly contributed to the establishment of this theory. IEs appear in many applications ranging from physics and chemistry, to biology and engineering \cite{Waz^2,groetsch2007integral}, for instance in potential theory, diffraction and inverse problems such as scattering in quantum mechanics \cite{Waz^2,groetsch2007integral,Lak}. Neural field equations, that model brain activity, can be described using IEs and integro-differential equations (IDEs), due to their highly non-local nature (\cite{amari1977dynamics}). IEs are related to the theory of ordinary differential equations (ODEs) and partial differential equations (PDEs), however they possess unique properties. While ODEs and PDEs describe local behavior, IEs model global (long-distance) spatiotemporal relations. Moreover, ODEs and PDEs have IE forms that in certain circumstances can be solved more effectively and efficiently due to the better stability properties of IE solvers compared to ODE and PDE solvers \cite{rokhlin1985rapid,rokhlin1990rapid}. See also \cite{greengard1998integral} for an example of PDE system that is solved with high accuracy through an IE method. 

Learning nonlocal operators for dynamics with long-distance relations is an open problem in deep learning.
In this article, we introduce and address the problem of learning nonlocal dynamics from data through integral equations. Namely, we introduce the {\it Neural Integral Equation} (NIE) and the {\it Attentional Neural Integral Equation} (ANIE). Our setup is that of an operator learning problem, where we learn the integral operator that generates dynamics that fit given data. Often, one has observations of a dynamical system without knowing its analytical form. Our approach permits modeling the system purely from the observations. This model, via the learned integral operator, can be used to generate dynamics, as well as be used to infer the spatiotemporal relations that generated the data. The innovation of our proposed method lies in the fact that we formulate the operator learning problem associated to dynamics in the form of an optimization problem for the solutions of an IE obtained through an IE solver. Unlike other operator learning methods that learn dynamics as a mapping between function spaces for fixed time points, i.e. as a mapping $T : \prod_i \mathcal A_i \longrightarrow \prod_j \mathcal B_j$, where $\mathcal A_i$ and $\mathcal B_j$ are function spaces each representing a time coordinate, NIE and ANIE allow to continuously learn dynamics with arbitrary time resolution. Our solver outputs solutions through an iterative procedure \cite{Waz^2}, which converges to a solution of the IE.

\begin{figure*}[ht]
\centering
  \includegraphics[width=0.8\textwidth]{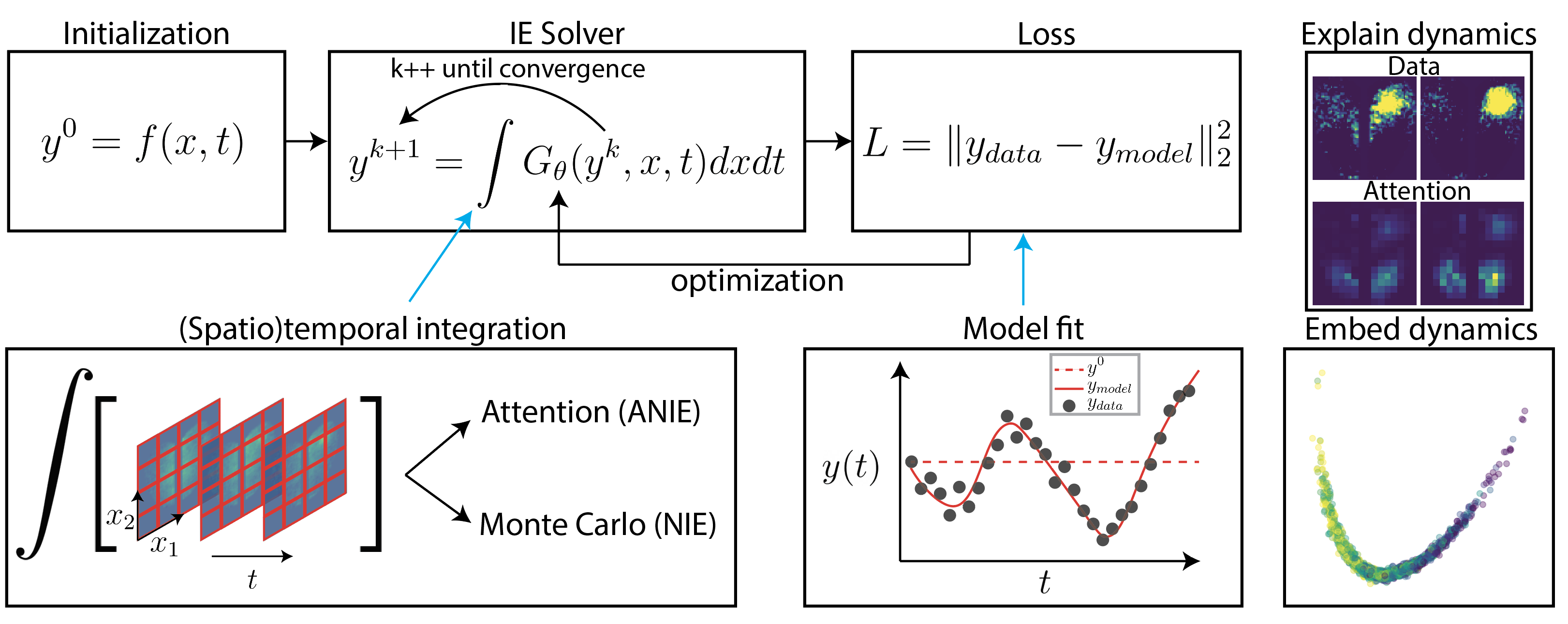}
  \caption{Diagrammatic representation of the model. The solver is initialized with $f$, also called the free function. This initialization is often the first time point of the dynamics. To solve the IE and find the solution $\mathbf y$, an iterative procedure is carried out in which at each solver step $k$ the integral of $G_\theta(\mathbf y^k,x,t)$ is computed and used as the solution $\mathbf y^{k+1}$ in the next step. Integration is done either with Monte Carlo (via torchquad) or with self-attention, representing NIE and ANIE respectively. The solver integration steps are repeated until convergence of $\mathbf y^k$ to the solution of the IE. This solution is then compared to the input data to compute a loss that, via backpropagation, is used to find $\theta$ that minimizes the error. The resulting integral operator represents the IE that models the data. Top right, an example of the attention weights for calcium imaging dynamics is presented. Bottom right, an example of dynamimcal embedding of Navier-Stokes dataset colored by velocity is shown.}\label{fig:Model}
\end{figure*}

\subsection{Our Contributions}

In this article, we introduce Neural Integral Equations (NIE) and Attentional Neural Integral Equations (ANIE), which are novel neural network based methods for learning dynamics, in the form of IEs, from data. Our architectures allow modeling dynamics with long-distance spatiotemporal relations typical of 
non-local functional equations. Our main contributions are as follows:

\begin{itemize}
    \item 
    We introduce a novel method for learning dynamics from data as solutions of IEs of the second kind through an IE solver.
    \item 
    We implement a fully differentiable IE solver in PyTorch\footnote{Code available at: \url{https://github.com/emazap7/ANIE}}.
    \item 
    We implement a highly scalable version of the solver where integration is done with a self-attention mechanism.
    \item 
    We derive theoretical results on convergence of the solver, and approximation capabilities of our models.
    \item 
    Our model provides explainable dynamics and meaningful embeddings of these dynamics.
    \item 
    Finally, we use our method to model and interpret non-local dynamics from brain activity recordings.
\end{itemize}

\subsection{Background and related work}

\subsubsection{Integral equations in numerical analysis}

Due to their wide range of applications, the theory of IEs has attracted the attention of mathematicians, physicists, and engineers for a long time. Detailed accounts on integral equations can be found in \cite{Zem,Waz^2,bocher1926introduction}. Along with their theoretical properties, much attention has been devoted to the development of efficient integral equation solvers, focusing on rapidly obtaining highly accurate solutions of certain PDE systems \cite{rokhlin1985rapid,rokhlin1990rapid}. In fact, it is known that integral equation solvers obtain more accurate solutions than differential solvers for a variety of ODEs and PDEs. 

\subsubsection{Operator learning}

IE solvers are used to solve given equations through some iterative procedure, see e.g. \cite{Waz^2,delves1988computational}. Moreover, machine learning approaches to solve given types of IEs have been implemented \cite{guan2022solving,que2016integral,keller2019integral,guo2021solving,effati2012neural}. In such cases, the IE is known, and we seek the solution of it. However, in practice, we often do not have access to the analytical form of the equation and we only have data sampled from a system. In such cases, we want to model the system by learning an operator that can reproduce the system. This is the setting of operator learning problems, and several approaches to operator learning, including using deep learning, have been presented \cite{kovachki2021neural,lu2021learning,li2020fourier,li2020neural,cao2021choose,hao2023gnot,maier2022known,kovachki2024operator,poli2022transform,bartolucci2024representation,ovadia2023realtime,oommen2022learning}. Typical operator learning problems are formulated on finite grids (finite difference methods) that approximate the domain of the functions. In this case, recovering the continuous limit is a very challenging problem, and irregularly sampled data can completely alter the evaluation of the learned operator. Operator learning for integral equations has not been considered thus far, and it constitutes the main novelty of the present article. This is entailed in the formulation of the operator learning problem through an IE solver. The convenience of this approach lies in the capability of the solver to sample the domain of integration continuously, and the capabilities of integral equations to model very complex dynamics, due to their highly non-local behavior. A similar approach for Integro-Differential Equations (IDEs) has been followed in \cite{NIDE}. However, in the present work our implementation does not include differential solvers, and the reformulation of such dynamical problems in terms of IEs has great benefits in terms of solver speed and stability. Moreover, our version of an IE solver that approximates integrals via self-attention allows for higher dimensional integrals than those considered in \cite{NIDE}.

\subsubsection{Learning continuous dynamics}

Modeling continuous dynamics from discretely sampled data is a fundamental task in data science. Methods for continuous modeling include those based on ODEs \cite{NODE,NODE2}. While ODEs are useful for modeling temporal dynamics, they are fundamentally local equations which neither model spatial nor long-range temporal relations. The authors of \cite{NODE} have employed auxiliary tools, such as RNNs, in order to include non-locality. We point out that RNNs can be seen as performing a temporal integration (in discrete steps), in order to codify some degree of non-local (temporal) dependence in the dynamics. In this work, we introduce a framework that provides a more general and formal solution to this non-local integration problem. Moreover, the dynamics are not produced sequentially with respect to time, as is done by ODE solvers, but are processed in parallel thus providing increased efficiency, as we will demonstrate experimentally. 

\subsubsection{Integration via self-attention}

The self-attention mechanism and transformers were introduced in \cite{vaswani2017attention} and applied to machine translation tasks. Thanks to their initial success, they have since been used in many other domains, including operator learning for dynamics \cite{cao2021choose,geneva2022transformers}. Interestingly, the self-attention mechanism can be interpreted as the Nystrom method for approximating integrals \cite{xiong2021nystromformer}. Making use of this connection, we approximate the integral kernel of our model using self-attention, allowing efficient integration over higher dimensions.

\section{Neural integral equations}\label{sec:NIE}

An IE (of Urysohn type) takes the general form given by
\begin{equation}\label{eqn:gen_NIE}
    \bold y(t) = f(t) + \int_{\alpha(t)}^{\beta(t)} G(\bold y(s),t,s) ds,
\end{equation}
where the variable $s$ is the local time used for integration for each $t$, which is the global time. Due to their fundamentally non-local behavior, integral equations have been used to model physical and biological phenomena, such as brain dynamics, virus spreading, and plasma physics \cite{Waz^2,groetsch2007integral,amari1977dynamics}. 
The case considered in this article, where the indeterminate function $\bold y(t)$ appears both under the sign of integration and outside of it, is termed an equation of the second kind, as opposed to the first kind where the indeterminate function appears only in the integral operator. IEs of the second kind are more stable than of the first kind for reasons rooted in functional analysis, as explained in Appendix~\ref{sec:existence}.

We introduce Neural Integral Equations (NIEs), a deep neural network model based on integral equations. NIEs are integral equations as defined by Equation~\ref{eqn:gen_NIE} where $G$ is a neural network, parameterized by $\theta$, and indicated by $G_\theta$. Training a NIE consists of optimizing $G_\theta$ in such a way that the corresponding solution $\bold y$ to Equation~\ref{eqn:gen_NIE} fits the given data. At each step of the training, we perform two fundamental procedures. The first one is to solve the IE determined by $G_\theta$, and the second one is to optimize for $G_\theta$ in such a way that solving the corresponding IE produces a function that fits a given dataset. Details on the solver procedure and the training are given in Appendix~\ref{sec:solver}.

IEs, in contrast to ODEs and PDEs, are non-local equations \cite{stech1981integral} since in order to evaluate the integral operator $\int_{\alpha(t)}^{\beta(t)} G_\theta( \bullet ,t,s) ds : \mathcal A \longrightarrow \mathcal A$ on a function $\mathbf y$, we need the value of $\mathbf y$ over the full integration domain. In fact, to evaluate 
the RHS of Equation~\ref{eqn:gen_NIE} at an arbitrary time point $t$ the function $\bold y(s)$ between $\alpha(t)$ and $\beta(t)$ is needed. Here, $\alpha$ and $\beta$ are arbitrary functions and common choices include $\alpha(t) =a$ and $\beta(t) = b$ (called Fredholm equations), or $\alpha(t) = 0$ and $\beta(t) = t$ (called Volterra equations). Consequently, solving an integral equation requires an iterative procedure, based on the notion of Picard iterations (successive approximation method), where the solution is obtained as a sequence of approximations that converge to the solution. We refer the reader to Appendix~\ref{sec:solving} for details on the solver implemented in this article, the theory upon which it is based, and the proofs regarding the convergence of our algorithms to a solution of the given IE (see Theorem~\ref{thm:iteration_method} and Corollary~\ref{cor:iterative_method}). We also refer to \cite{Waz^2} for an elementary and computationally driven introduction to the theory behind the methods that motivate this procedure, and \cite{delves1988computational} for a more detailed account. 

Interestingly, utilizing NIEs to model ODEs allows to bypass the use of the ODE solvers, as the one introduced in \cite{NODE,NODE2}. The convenience in this approach is that the integral equation solver is more stable than the ODE solver \cite{kushnir2012highly}. ODE solver instabilities, induced by equation stiffness, have been previously considered in \cite{STEER,finlay2020train}. The IE solver presented in this work thus does not suffer from these problems, and is also significantly faster. 

\begin{figure*}[ht]
\centering
  \includegraphics[width=0.8\textwidth]{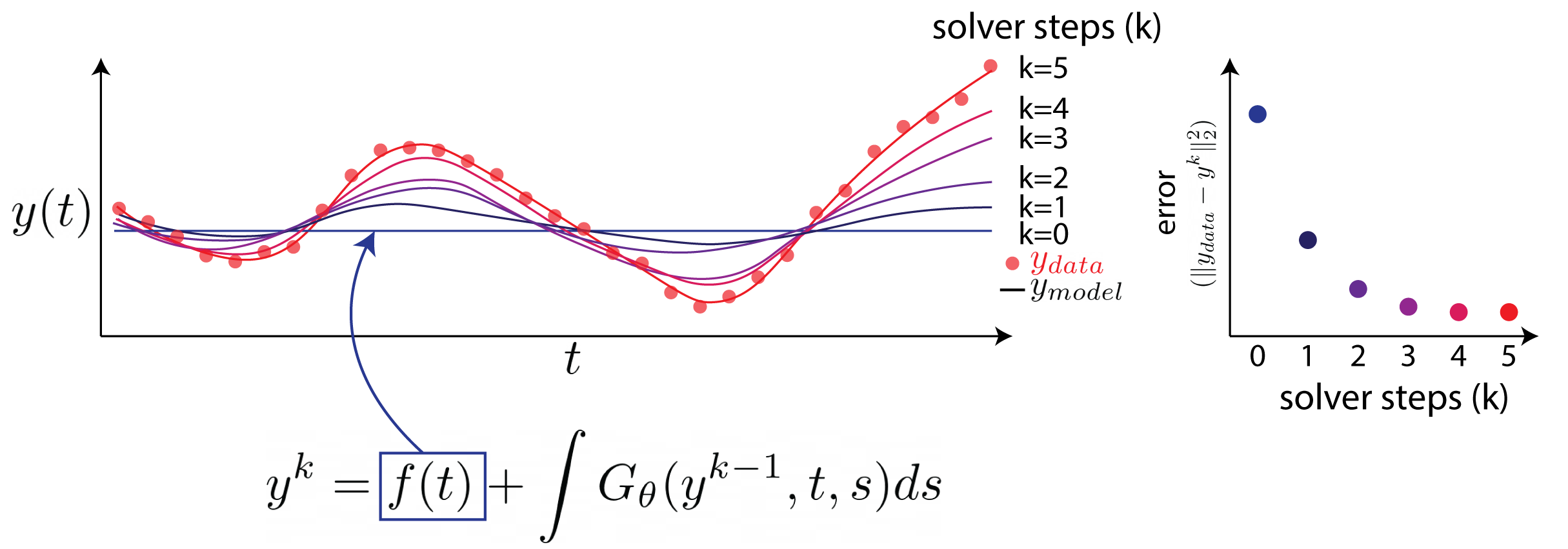}
  \caption{Diagrammatic representation of the IE solver procedure. The solver is initialized with the free function $\mathbf y^0 := f$. The integral operator is applied to $\mathbf y^0$, and a new guess $\mathbf y^1$ is obtained. This is repeated until convergence to a solution. The left panel shows the solution as a function of solver steps. The right panel shows the error of the solution as a function of solver steps.}\label{fig:Solver}
\end{figure*}

It is often useful to consider a more specific form for IEs, where the function $G$ factors in the product of a {\it kernel} $K$ and a generally non-linear function $F$ as $G(\mathbf y,t,s) = K(t,s)F(\mathbf y)$. Here, $K$ is matrix-valued and it carries the dependence on the time (both $t$ and $s$), while $F$ depends only on the indeterminate function $\mathbf y$. The form of this IE is therefore:
\begin{equation}\label{eqn:NIE}
    \bold y(t) = f(t) + \int_{\alpha(t)}^{\beta(t)} K(t,s)F(\bold y(s))ds.
\end{equation}
NIEs in this form comprise two neural networks, namely $K$ and $F$.
We observe that in IEs, the initial condition is embedded in the equation itself, and it is not an arbitrary value to be specified as an extra condition. To solve the IE we implement a solver that performs an iterative procedure to obtain a solution, see Appendix \ref{sec:solving} and Appendix \ref{sec:solver}. During the iterations, Monte Carlo sampling is performed to evaluate the integrals. This procedure allows our deep learning model to be independent of the temporal grid points, therefore resulting in a continuous model, since the model internally uses randomly sampled points to generate the successive iterations, as opposed to using fixed grid points. The general algorithm for training NIE is given in Algorithm~\ref{algo:NIE}, and a diagrammatic overview of it is shown in Figure~\ref{fig:Model}. See also Figure~\ref{fig:Solver} for a visualization of the general solving procedure. 

\begin{algorithm}
    \caption{NIE method training step. Integration is performed using the module torch.quad, with the Monte Carlo method.}
    \label{algo:NIE}
    \begin{algorithmic}[1]
        \Require{$\mathbf y_0(t)$} \Comment{Initialization}
        \Ensure{$\mathbf y(t)$} \Comment{Solution to IE with initial $\mathbf y_0(t)$}
        \State{$\mathbf y^0(t) := \mathbf y_0(t)$} \Comment{Initial solution guess}
        \While{${\rm iter}\leq {\rm max iter}$ and ${\rm error} > {\rm tolerance}$}
        \State{Evaluate: $\mathbf y^{i+1}(t) = f(\mathbf y^i,t) + \int_{\alpha(t)}^{\beta(t)}G(t,s,\mathbf y^i(s))ds$}
        \State{Set solution to be: $r \mathbf y^{i} + (1-r)\mathbf y^{i+1}$}
        \State{New error: ${\rm error} = {\rm metric}(\mathbf y^{i+1},\mathbf y^i$)}
        \EndWhile
        \State{Output of solver: $\mathbf y(t)$}
        \State{Compute loss wrt observations: ${\rm loss}(\mathbf y(t),{\rm obs})$}
        \State{Gradient descent step}
    \end{algorithmic}  
\end{algorithm}


\subsection{Space, time and higher dimensional integration}

IEs can have multiple space dimensions in addition to time. Such equations are formulated as 
\begin{equation}\label{eqn:PIE}
    \bold y(\bold x,t) = f(\bold x,t) + \int_{\Omega}\int_{\alpha(t)}^{\beta(t)} G(\bold y(\bold x',s), \bold x, \bold x', t, s) d\bold x' ds,
\end{equation}
where $\Omega \subset \mathbb R^n$ is a domain in $\mathbb R^n$, and $\bold y: \Omega \times I \longrightarrow \mathbb R^m$, for some interval $I\subset \mathbb R$.
More commonly, in the literature, one finds a simpler case of higher dimensional IEs, where the integral component $\int_{\Omega}\int_{\alpha(t)}^{\beta(t)} G(\bold y(x',s), \bold x, \bold x', t, s) d\bold x' ds$ is obtained as a sum of terms with only partial integrations. Such an equation takes the form
\begin{eqnarray}\label{eqn:PIE2}
\begin{aligned}
    \bold y(\bold x,t) &=& f(\bold x,t) + \int_{\alpha(t)}^{\beta(t)} G_1(\bold y(\bold x,s), \bold x, t, s) ds + \int_{\Omega} G_2(\bold y(\bold x',t), \bold x, \bold x', t) d\bold x',
\end{aligned}    
\end{eqnarray}
These equations are the integral counterpart of PDEs, similarly to the relation between one-dimensional IEs and ODEs, and they are called Partial Integral Equations (PIEs). With slight abuse of notation we will still refer to Equation~\ref{eqn:PIE} as a PIE, as we will in practice use such an approach to model PDEs in the case of Burgers' equation and the Navier-Stokes equation.

\subsection{Attentional neural integral equations}\label{sec:ANIE}

Training of NIE requires an integration step at each time point, incurring in a potentially high computational cost. This integration step is implemented using the torchquad package \cite{gomez2021torchquad}, a high performance numerical Monte Carlo integration method, resulting in fast integration and high scalability of NIE. For example, solving ODEs using NIE is significantly faster than using traditional ODE solvers (Table \ref{tab:walltime}). However, several limitations are associated with the torchquad integration method. In fact, torchquad requires significantly increasing numbers of sampled points with increasing numbers of dimensions. To use NIE for solving PDEs and (P)IEs we require efficient spatial integration in high dimensions.

To  address these challenges, we have employed an approach to NIE where the integral operator is based on a self-attention mechanism. In fact, self-attention can be viewed as an approximation of an integration procedure \cite{tsai2019transformer,xiong2021nystromformer}, where the product of queries and keys coincides with the notion of a kernel, as the one discussed in Section~\ref{sec:NIE}. In \cite{cao2021choose} the parallelism between self-attention and integration of kernels was further explored to interpret transformers as Galerkin projections in operator learning tasks.

We have replaced the analytical integral $\int_{\alpha(t)}^{\beta(t)} G(t,s,\bold y(s))ds$ in Equation~\ref{eqn:gen_NIE} with a self-attention procedure. The resulting model, which we call {\it Attentional Neural Integral Equation} (ANIE), follows the same principle of iterative IE solving presented in Section~\ref{sec:NIE} but where the neural networks $K$ and $F$ are replaced by attention matrices. It can be shown, see Appendix~\ref{sec:solving}, that the successive approximation method is still applicable in this case to obtain a solution for the corresponding equation. Observe, following the comparison between integration and self-attention, that $K$ is decomposed in the product of queries and keys, as described for instance in \cite{cao2021choose}. The interval of integration $[\alpha(t),\beta(t)]$ is determined, in the attentional approximation, by means of the mask. In particular, if there is no mask we have a Fredholm IE, while the causal attention mask \cite{yang2021causal} corresponds to a Volterra type of IE.

An iterative procedure similar to the one discussed in Algorithm~\ref{algo:NIE} is implemented to solve the corresponding IE, see Appendix \ref{sec:solving} and Appendix \ref{sec:ANIE_solving_training}. During iterations, we sample points uniformly from the spatiotemporal domain, and the corresponding integral operator does not depend on the grid points of the dataset. Our experiments on the Burgers' dataset (Section~\ref{sec:PDE} show that our model is stable with respect to change of spatiotemporal stamps since the model internally uses randomly sampled points to generate the successive iterations, rather than the fixed grid points. A detailed description of the integration procedure, along with solver steps and training for ANIE is given in Appendix~\ref{sec:ANIE_solving_training}. Moreover, Theorem~\ref{thm:iteration_method}, Corollary~\ref{cor:iterative_method} and Remark~\ref{rmk:iterative_method} show that the solver procedure converges to a solution under certain mild assumptions. 

Algorithm~\ref{algo:ANIE} summarizes the solving and training procedures for ANIE. A detailed description of the meaning of $\frak{Att}$ is found in Appendix~\ref{sec:ANIE_solving_training}. Theoretical considerations on Fredholm generalized equations with general operators, integral operator approximation through self-attention, and existence of the solutions for these equations are given in Appendix~\ref{sec:existence}. Figure~\ref{fig:Transformer_integration} gives a diagrammatic representation of the integration procedure implemented in this article, and Figure~\ref{fig:spatial_IE} gives a schematic representation of the solver procedure with space and time.

\begin{algorithm}
    \caption{ANIE method training step. Integration here is replaced by a transformer employing self-attention.}
    \label{algo:ANIE}
    \begin{algorithmic}[1]
        \Require{$\mathbf y_0(\mathbf x, t)$} \Comment{Initialization}
        \Ensure{$\mathbf y(\mathbf x, t)$} \Comment{Solution to IE with initial $\mathbf y_0(\mathbf x, t)$}
        \State{$\mathbf y^0(\mathbf x, t) := \mathbf y_0(\mathbf x, t)$} \Comment{Initial solution guess}
        \While{${\rm iter}\leq {\rm max iter}$ and ${\rm error} > {\rm tolerance}$}
        \State{Concatenate space and time tokens to $\mathbf y^i(\mathbf x,t)$: $\tilde{\mathbf y}^i(\mathbf x,t) = {\rm concat}(\mathbf y^i(\mathbf x,t),s,t)$}
        \State{Evaluate with self-attention: $y^{i+1}(t) = f(\tilde{\mathbf y}^i,t) + {\mathfrak Att}(\tilde{\mathbf y}^i(\mathbf x,t))$}
        \State{Set solution to be: $r \mathbf y^{i} + (1-r)\mathbf y^{i+1}$}
        \State{New error: ${\rm error} = {\rm metric}(\mathbf y^{i+1},\mathbf y^i$)}
        \EndWhile
        \State{Output of solver: $\mathbf y(\mathbf x, t)$}
        \State{Compute loss wrt observations: ${\rm loss}(\mathbf y(\mathbf x, t),{\rm obs})$}
        \State{Gradient descent step}
    \end{algorithmic}  
\end{algorithm}

\section{Experiments}\label{sec:experiments}

\subsection{Modeling PDEs with IEs: Burgers' and Navier-Stokes equations}\label{sec:PDE}

PDEs can be reformulated as IEs in several circumstances, and dynamics generated by differential operators can therefore be modeled through ANIE as a PIE, where integration is performed in space and time. We consider two well known types of PDEs, namely the Burgers' equation and the Navier-Stokes equation. Since NIE is implemented only for time integration, we use only ANIE in these experiments, which allows for efficient space and time integration. We observe that our implementation of Algorithm~\ref{algo:ANIE} applied to the case of Navier-Stokes equation closely parallels the IE method employed in \cite{greengard1998integral}, with the main difference that we learn the Green's function through gradient descent, since no knowledge of the underlying Navier-Stokes equations is assumed. 

\begin{table}[ht]
\caption{Left, benchmark on the Navier-Stokes equation. We evaluate the models on predicting dynamics of different lengths ($t=3, 5, 10, 20$) for unseen initial conditions. The models that use a single time point are ANIE (ours), FNO2D, ViT~\cite{dosovitskiy2020image}, ViTsmall~\cite{lee2021vision} and ViTparallel~\cite{touvron2022three} models, while the convolutional LSTM, FNO3D, and ViT3D all use more time points ($2$, $10$ and $2$, respectively) to predict the rest of the dynamics. ANIE outperforms also the models that use more data points for initialization. Right, benchmark on the Burgers' equation with different time intervals $t=10, 15, 25$ and space resolutions $s=256, 512$, where a time interpolation task is performed. A symbol $-$ has been used to indicate those models that were not suitable for certain experiments (e.g. wrong dimensionality), while NA indicates models that did not converge, or did not fit in memory.}\label{tab:NS_Bur} 
\centering

\resizebox{\textwidth}{!}{\begin{tabular}{c c c c c | c c c c c c }
\toprule
& \multicolumn{4}{c}{Navier-Stokes} & \multicolumn{6}{c}{Burgers'}\\
\midrule 
 & t = 3 & t = 5 & t = 10 & t = 20 & \multicolumn{2}{c}{t=10} & \multicolumn{2}{c}{t=15} & \multicolumn{2}{c}{t=25}\\
 \hline
 &  &  &  &  & s = 256 & s = 512 & s = 256 & s = 512 & s = 256 & s = 512 \\
   \midrule
  LSTM & $.1384$ & $.2337$ & $.1422$ & $.2465$ & $-$ & $-$ & $-$ & $-$ & $-$ & $-$\\
  ResNet & $-$ & $-$ & $-$ & $-$ & $.0295$ & $.0309$ & $.0280$ & $.0232$ & $.0194$ & $.0204$\\
  Conv1DLSTM & $-$ & $-$ & $-$ & $-$  & $.0132$ & $.0133$ & $.0132$ & $.0136$ & $.0124$ & $.0134$\\
  Conv2DLSTM & $.4935$ & $.4393$ & $.3931$ & $.2999$ & $-$ & $-$ & $-$ & $-$ & $-$ & $-$\\
  FNO1D & $-$ & $-$ & $-$ & $-$  & $.0088$ & $.0088$ & $0087$ & $.0087$ & $.0083$ & $.0086$\\
  Galerkin & $-$ & $-$ & $-$ & $-$ & $.0525$ & NA & $.0521$ & NA & $.0518$ & NA\\
  FNO2D & $.2795$ & $.2724$ & NA & NA & $-$ & $-$ & $-$ & $-$ & $-$ & $-$\\
  FNO3D & NA& NA & $.1751$ & $.0701$ & $-$ & $-$ & $-$ & $-$ & $-$ & $-$\\
  ViT & $.1093$ & $.0877$& $.2473$& $.2367$ & $.0430$ & $.0423$ & $.0423$ & $.0422$ & $.0422$ & $.0424$ \\
  ViTsmall & $.0926$ & $.0702$ & $.0677$ & $.0655$ & $.0429$ & $.0429$ & $.0426$ & $.0427$ & $.0417$ & $.0424$\\
   ViTparallel & $.2901$ & $.2660$ & $.2475$ & $.2368$ & $.0433$ & $.0702$ & $.0573$ & $.0861$ & $.0435$ & $.0700$\\
  ViT3D & $.0360$& $.0365$& $.0433$& $.0406$ & $-$ & $-$ & $-$ & $-$ & $-$ & $-$ \\
  \textbf{ANIE (ours)} & $\mathbf{.0194}$ & $\mathbf{.0220}$ & $\mathbf{.0193}$ & $\mathbf{.0117}$ & $\mathbf{.0024}$ & $\mathbf{.0026}$ & $\mathbf{.0024}$ & $\mathbf{.0024}$ & $\mathbf{.0022}$ & $\mathbf{.0023}$\\
\bottomrule
\end{tabular}}\\
\end{table}\

For the Burgers' equation, we focus on the ability of ANIE to model both space and time continuously, and we therefore perform an interpolation taks, where the model outputs time points that are not included in the training test, and for unseen initial conditions. This is in contrast to \cite{cao2021choose,li2020fourier} where a ``static'' Burgers' equation was considered in which the learned operator maps the initial condition ($t = 0$) to the final time ($t=1$), thus treating time as a discrete two point set. In our approach, we model the system continuously over a time interval and randomly sample points during iterations to perform the quadrature of the temporal integrals. In this experiment, the Galerkin model \cite{cao2021choose}, was not included for the higher spatial dimension setting because the amount of memory required exceeded what was available to us during the experiments. The results are reported in Table~\ref{tab:NS_Bur} (right side), and an example of the learned dynamics is given in Figure~\ref{fig:Burgers}.

\begin{figure*}[ht]
\centering
  \includegraphics[width=0.8\textwidth]{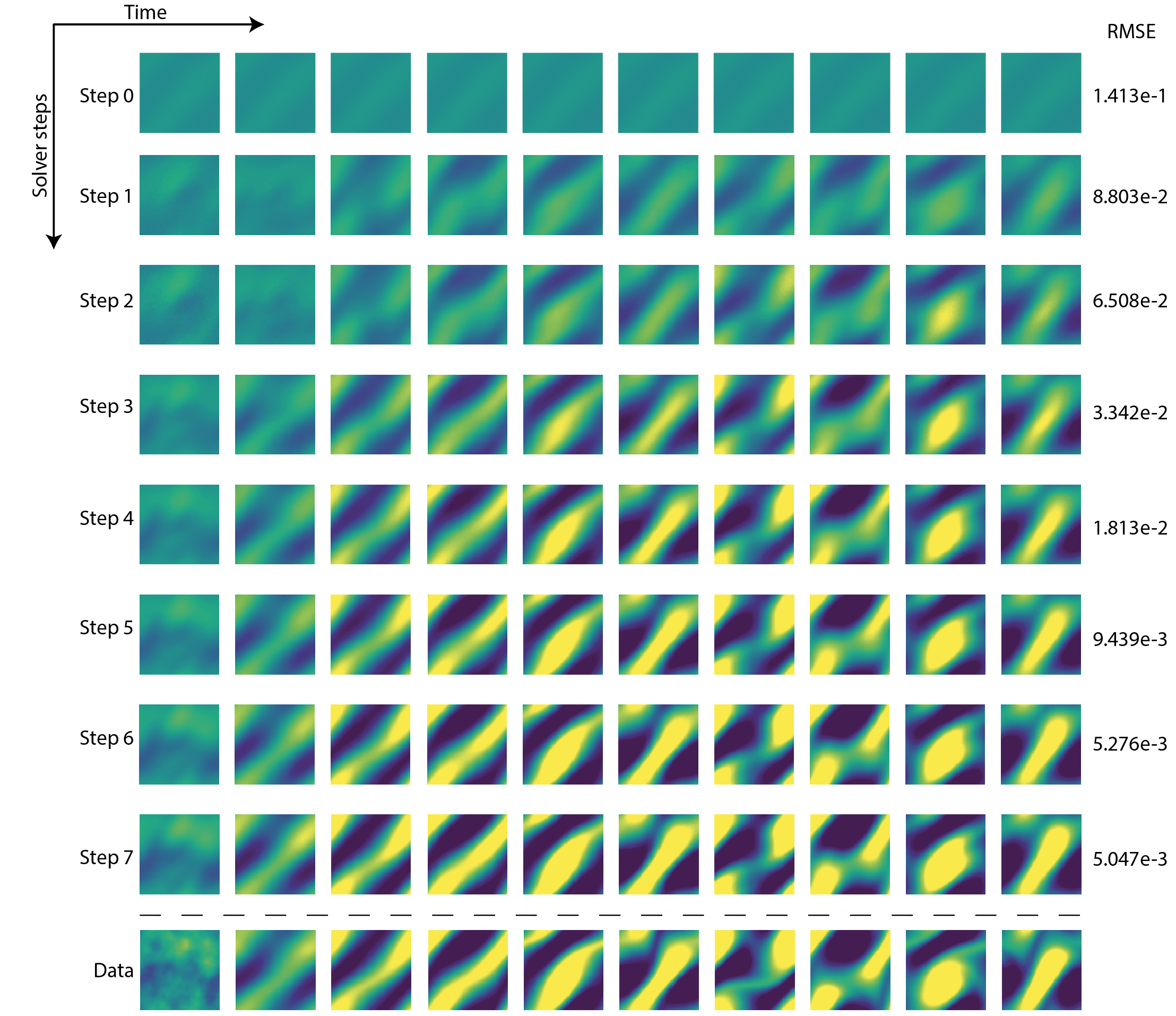}\caption{
  Example dynamics of the (2+1)D Navier-Stokes system, where the model is initialized only with the first frame of the dynamics. Ground truth data is given at the bottom. Along with the final prediction (Step 7), the subsequent solver guesses are shown. The error during the solution generation are reported on the right. The figure shows also that the solver converges when producing the final output, cf. Figure~\ref{fig:iter_conv}. 
  }\label{fig:iter_error}
\end{figure*}

For the Navier-Stokes equation, we consider an extrapolation task where we evaluate the model on unseen initial conditions. Previous works have shown high performance in predicting dynamics of Navier-Stokes from new initial conditions, but they require several frames (i.e. several time points) to be fed into the model in order to achieve such performance. We see that since ANIE learns the full dynamics from arbitrarily chosen initial conditions, we achieve good performance even when a single initial condition is used to initialize the system. We train FNO2D, ViT, ViTsmall and ViTparallel with initialization on a single time point, while  convolutional LSTM, FNO3D, and ViT3D are trained with $2$, $2$ and $10$ times for initialization, respectively. The results are given in Table~\ref{tab:NS_Bur} (left side).
We note that ANIE outperforms also the models that use more data points for initialization. FNO2D did not converge for higher number of points, and therefore results for time points $t=10$ and $t=20$ have not been reported, while for FNO3D, we have conducted the experiments only for $t=10$ and $t=20$ since using fewer points for the time dimension would have effectively reduced FNO3D to FNO2D. An example of the Navier-Stokes dynamics is given in Figure~\ref{fig:iter_error}.


\subsection{Modeling brain dynamics using ANIE}\label{sec:Brain_dyn}

Brain activity can be modeled as a spatiotemporal dynamical system \cite{vyas2020computation}. While most connections between neurons are localized in space, there are a significant number of interactions that are long-range \cite{ercsey2013predictive}. As such, brain dynamics can be modeled using integral equations \cite{amari1977dynamics} which, unlike PDEs, allow for non-local interactions. Since ANIE allows efficient learning of integral operators from data, we demonstrate the ability of ANIE to learn non-local brain dynamics from functional Magnetic Resonance Imaging (fMRI) recordings.

To obtain fMRI data that has arbitrary time duration as well as unlimited trials, we make use of \textit{neurolib} \cite{cakan2021neurolib}, an fMRI simulation package. The data provided by this tool permit for more extensive comparison and statistical power. \textit{neurolib} simulates whole-brain activity using a system of delay differential equations, which are non-local equations, thus allowing testing of ANIE's ability to model non-local systems. Here we show the performance of ANIE and other models in modeling data generated by \textit{neurolib}. Details about data generation and preprocessing can be found in the Appendix~\ref{sec:fMRI_data}.

The generated fMRI data comprises neural activity for 80 nodes localized across the cortex. The first half of the data is used for training and the second half is used for testing. For training, the data are divided into segments of 20 time points, where the first time point is used as the initial condition, and the loss is computed over all 20 points. As such, the models are trained as an initial condition problem. During inference, the models are given points from the test set as new initial conditions and asked to extrapolate for the following 19 points. The mean error per point for 200 new initial conditions is shown in Figure ~\ref{fig:generated_fmri_error_per_point} and summarized in Table ~\ref{tab:generated_fmri_error_per_point}. Figure~\ref{fig:fMRI_plots} shows the data and model per fMRI recording node over time. We show that ANIE has significantly better performance than other benchmarked methods for medium ($t=10$) and long ($t=20$) time step predictions, demonstrating its ability to model non-local dynamics. For shorter and more localized dynamics ($t=5$), FNO1D shows better performance, which can be explained by the fact that FNO1D outputs the average of the initial points provided as the prediction for the first 5 time steps.




\subsection{Interpretable dynamics}\label{sec:interpretability}

In addition to modeling and generating new dynamics, it is useful to get insight into the underlying process that generates the dynamics. For example, in neuroscience, a major goal is to understand how specific brain activity patterns give rise to cognition, learning, and behavior.
To explore the interpretability of ANIE we carry out two experiments. For the first experiment, we augment the spacetime integration domain with a CLS token \cite{devlin2018bert}, such that each dynamics is projected into a single vector. This vector can then be related to specific properties of the dynamics. Specifically, we embed these vectors for different Navier-Stokes dynamics and find that the resulting manifold (projected using PCA) has a highly nonrandom structure. This is in contrast to the projection of the raw data (see Figure~\ref{fig:smiley}). To further explore the resulting dynamics manifold, we color it by the velocities of the dynamics, a property that was not explicitly seen by the model during training. We find that the manifold highly correlates with velocity whereas the embedding of the raw data has no such correlation. To quantify this we compute the kNN regression error on the embeddings with respect to the velocities and find that the embedding obtained from ANIE has significantly lower error (see Table~\ref{tab:embedding_knn}).



For the second experiment, we inspect the attention weights of the model when predicting brain dynamics (calcium imaging, Appendix \ref{appendix:calcium_imaging}), in order to infer which cortical loci drive neuronal dynamics. Figure~\ref{fig:ANIE_attention} shows that the motor and visual cortices are the areas of the brain with the highest attention values. 
We note that the attention plots are not directly correlated with the brain activity inputs, suggesting that they point to new information about the data. To validate this, we compare the performance of predicting the visual stimulus, which was not explicitly provided to the model, from either the raw data or the attention values using a kNN-regressor (k=3) (see Appendix \ref{appendix:calcium_imaging}).
In Table \ref{tab:regression_visual_stim} we show that the attention weights significantly ($p=0.035$) outperform the raw data, thus demonstrating that ANIE can provide insights into the modeled dynamics.


\subsection{Further experiments}

In Appendix~\ref{sec:extra_exp} we include several more experiments regarding the training speed of ANIE showcasing that it is significantly faster than ODE solver based models, hyperparameter sensitivity of the model, and modeling of IE dynamics, along with further tables and figures on the experiments of Subsections~\ref{sec:PDE}, ~\ref{sec:Brain_dyn} and ~\ref{sec:interpretability}. In Appendix~\ref{sec:convergence} we have explored the convergence of the solver to fixed points of the corresponding IE.

\section{Limitations}

An important consideration relates to the theoretical guarantees for convergence of the solver as considered in Appendix~\ref{sec:solving}. In fact, enforcing contractivity of the integral operator might be needed for certain types of applications as seen in Appendix~\ref{sec:solving}. However, we mention that Lipschitz constraints (in particular contractivity) have been long studied in machine learning, including the case of transformers \cite{kim2021lipschitz,qi2023lipsformer}. We therefore expect that in cases when the solver is unstable, and convergence is problematic, one can enforce convergence using a Lipschitz constraint.

\section{Conclusions}

We have presented a neural network-based integral equation solver that can learn an integral operator and its associated dynamics from data, and capture non-local spatiotemporal interactions. We have demonstrated the ability of our method to learn ODE, PDE, and IE systems better than other dynamics models, in terms of better predictability as well as scalability. We have studied the approximation capabilities of NIE and ANIE, and obtained theoretical guarantees for the convergence of the IE solver, enhancing the understanding of this deep-learning approach. Moreover, we have shown that ANIE is interpretable and produces meaningful dynamics embeddings, as demonstrated in our applications to the Navier-Stokes equation and brain activity dynamics. Since IEs can be used to model many real-world systems, we anticipate that NIE and ANIE will have broad applications in science and engineering.

\section{Data availability statement}

We have included descriptions regarding how to reproduce all synthetic datasets, used in Figures~\ref{fig:Burgers},~\ref{fig:iter_error},~\ref{fig:fMRI_plots},~\ref{fig:NS_output} as well as Table~\ref{tab:NS_Bur}. All synthetic datasets are available at \url{https://github.com/emazap7/ANIE}.
Access to the datasets can be obtained also at \url{https://figshare.com/articles/dataset/IE_spirals/25606242} for integral equation spirals, \url{https://figshare.com/articles/dataset/Burgers_1k_t400/25606149} for Burgers data, \url{https://figshare.com/articles/dataset/Navier_Stokes_Dataset_mat/25606152} for Navier-Stokes data, and \url{https://figshare.com/articles/dataset/fMRI_data/25606272} for the simulated fMRI data. Lokta-Volterra and Lorentz system datasets can be generated using the notebooks found at \url{https://github.com/emazap7/ANIE}. The calcium imaging dataset is not available open source. We have included a detailed account of the techniques used in \cite{barson2020simultaneous,hamodi2020transverse} (see Appendix~\ref{appendix:calcium_imaging}), where information on how to obtain the dataset is included. 

\section{Code availability statement}

All codes are available at: \url{https://github.com/emazap7/ANIE}. Detailed installation descriptions are found thereby. Jupyter notebooks for training and testing of the models for the main experiments are provided in the repository. Pre-trained models are accessible directlty, and instructions on how to run the notebooks are added in the form of comments throughout the notebooks. The main codes for the models, along with the experiments, are found in the directory ``IE\_source''. 

\section{Author contributions}

EZ conceived the algorithmic framework, obtained the theoretical results, and contributed to the numerical experiments. AF and JOC contributed to the numerical experiments. AM, MH and JC provided the calcium imaging data. DvD led the study and conceived the algorithmic framework. EZ, AF, JOC and DvD contributed to writing the article.   



    



\bibliographystyle{alpha}
\bibliography{refs}

\appendix

\section{Additional Experiments}\label{sec:extra_exp}

\subsection{Additional figures and tables}\label{sec:additional_figs_tabs}

We hereby report additional figures and tables that complement the results given in Section~\ref{sec:experiments}.

The model DeepOnet + UNET in Table~\ref{tab:generated_fmri_error_per_point} is implemented similarly to \cite{diab2023u}. 

\begin{figure*}[h]
\centering
  \includegraphics[width=\textwidth]{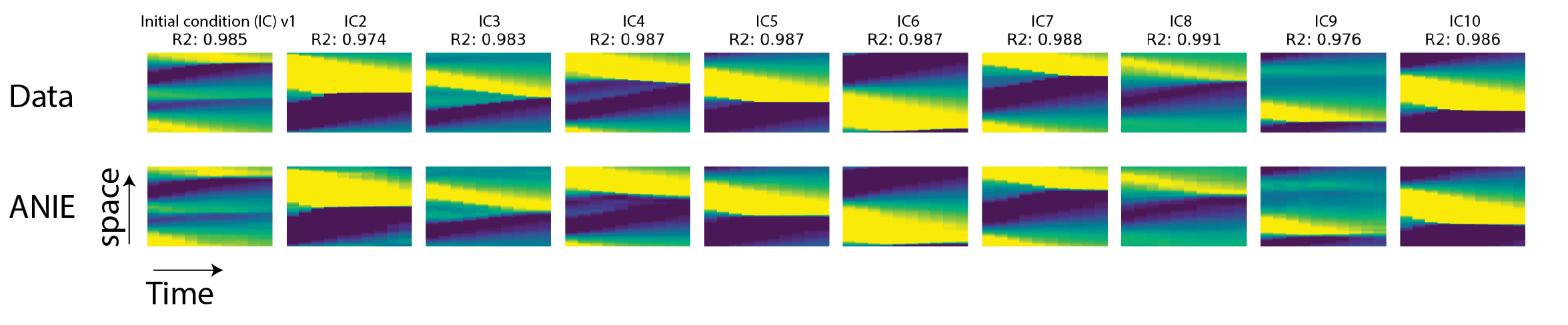}\caption{Example dynamics of the (1+1)D Burgers' equation. Each frame represents the full dynamics where the $\vec{x}$ axis shows time and the $\vec{y}$ axis shows space. Top row is data, and bottom row is ANIE prediction. Columns represent different dynamics resulting from different initial conditions. R2 values of model fits are shown for each of the dynamics.}\label{fig:Burgers}
\end{figure*}

\begin{table*}
\caption{Benchmark on predicting fMRI brain dynamics. We report the mean squared errors per extrapolated dynamics of different lengths ($t=5,10,20$) on new initial conditions. All models use a single data point as initial condition, while the LSTM model uses 2 time points. We see that as the dynamics gets more non-local (i.e. longer time intervals) only ANIE can correctly predict it, as shown by lower mean squared error.}\label{tab:generated_fmri_error_per_point}
\centering

\begin{tabular}{|c | c | c | c | c |}
  \hline 
  & t = 5 & t = 10 & t = 20\\
  \hline 
  NODE & $0.98 \pm 0.07831$ & $1.759	\pm 0.1407$ & $2.361 \pm 0.2227$\\
  \hline
  LSTM & $2.004 \pm 0.1856$ & $2.182 \pm 0.195$ & $2.47 \pm 0.1993$\\
  \hline
  Residual Network & $2.396 \pm 0.1705$ & $2.535 \pm 0.1706$ & $2.742 \pm 0.2$\\
  \hline
    FNO1D & \textbf{0.4735 $\pm$ 0.04857} & $1.5110 \pm 0.13570$ & $2.7320 \pm 0.31410$\\
  \hline
   ViT & $1.543 \pm 0.1235$ & $1.6650 \pm 0.1091$ & $2.0350 \pm 0.1497$\\
  \hline
   DeepOnet+AE & $2.436 \pm 0.5546$ & $2.774 \pm 1.018$ & $6.405 \pm 2.062$\\
   \hline
   DeepOnet+UNET & $2.692	\pm 0.5066$ & $3.065 \pm	0.962$ & $3.024 \pm 0.665$\\
  \hline
  \textbf{ANIE (ours)} & $0.7974	\pm 0.08118$ & $\mathbf{1.086 \pm 0.112}$ &  $\mathbf{1.242 \pm 0.1256}$\\
  \hline
\end{tabular}

\end{table*}

\begin{figure*}[h]
\centering
  \includegraphics[width=\textwidth]{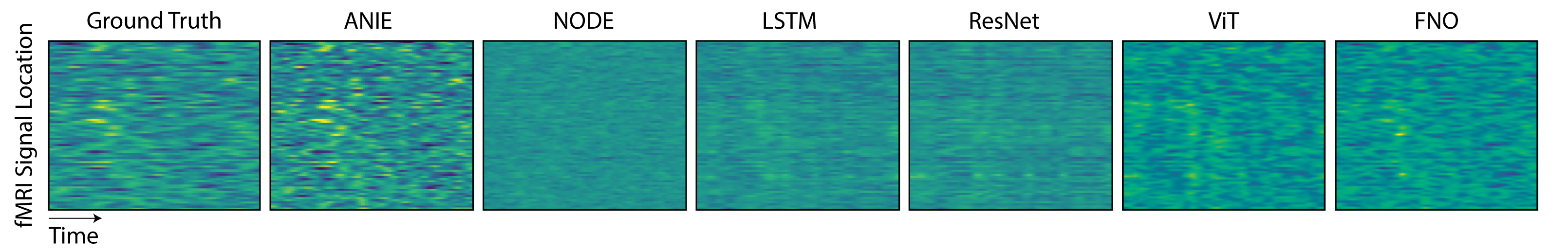}\caption{Example dynamics of fMRI data and corresponding model prediction. For each image, time is represented on the $\vec{x}$ axis, and cortical locations ($80$ nodes) are represented on the $\vec{y}$ axis.} 
  \label{fig:fMRI_plots}
\end{figure*}

\begin{figure*}[h]
\centering
  \includegraphics[width=0.8\textwidth]{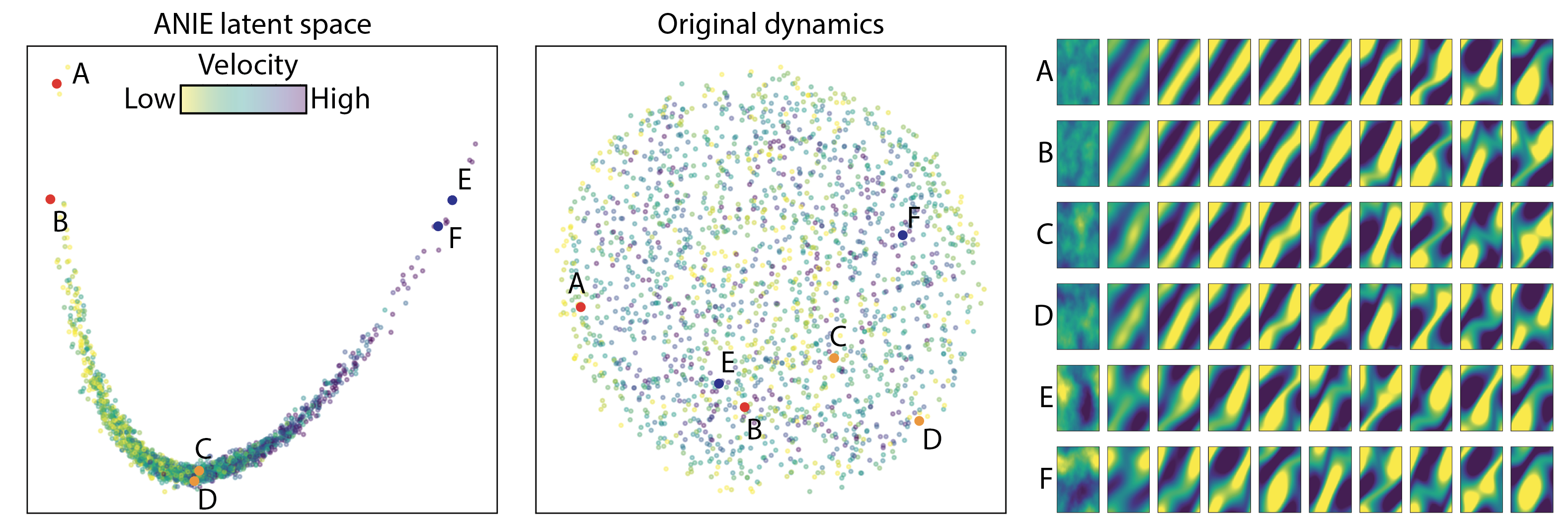}\caption{Embedding of Navier-Stokes dynamics using ANIE (Panel A), PCA (Panel B), and sample dynamics from the embedding spaces (Panel C). We see that the leftmost dynamics in Panel A correspond to lower velocity dynamics, and embedding smoothly transitions toward higher velocities from left to right. Such structure is lost when directly embedding using other methods (e.g. the reported PCA plot).}\label{fig:smiley}
\end{figure*}

\begin{figure*}[ht]
\centering
  \includegraphics[width=0.8\textwidth]{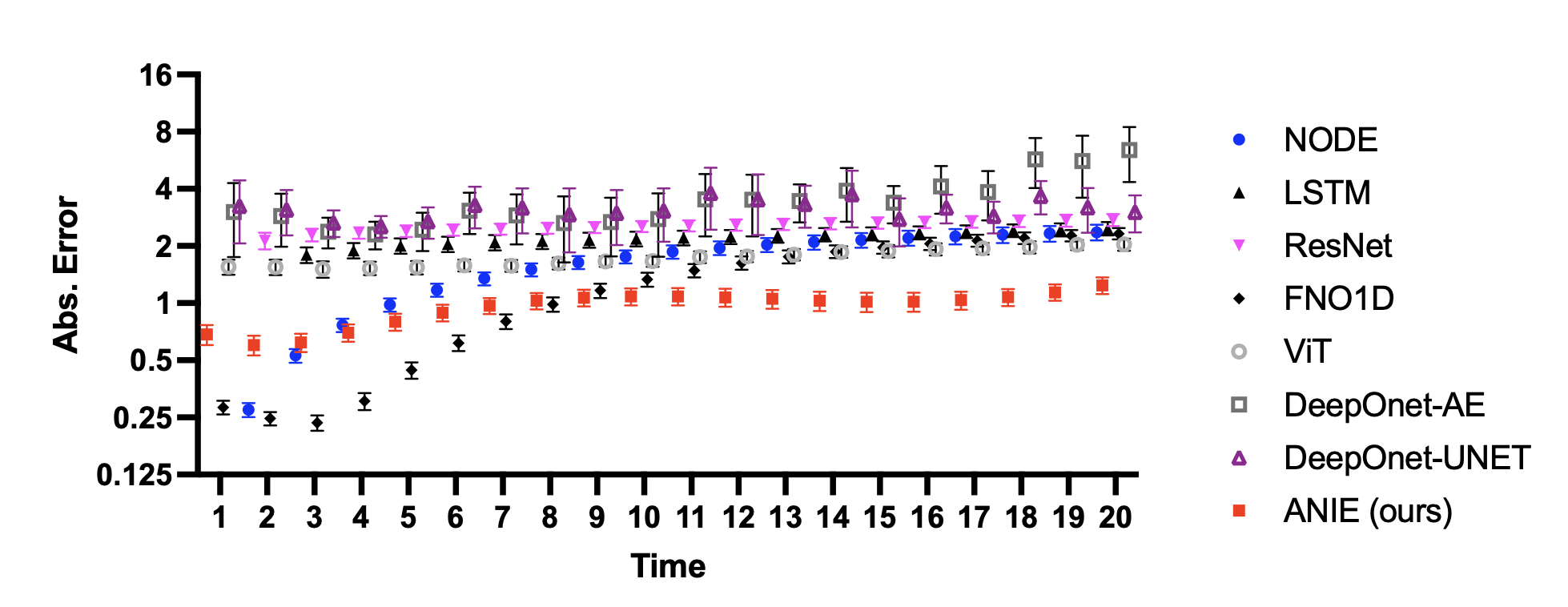}\caption{Quantification, using absolute error per time point, of model fits to simulated fMRI dataset. Models were run during inference on initial conditions not seen during training. ANIE has the best performance (lowest error) in predicting longer dynamics, which encompass a higher non-local component.}\label{fig:generated_fmri_error_per_point}
\end{figure*}

\begin{figure*}[h]
\centering
  \includegraphics[width=0.9\textwidth]{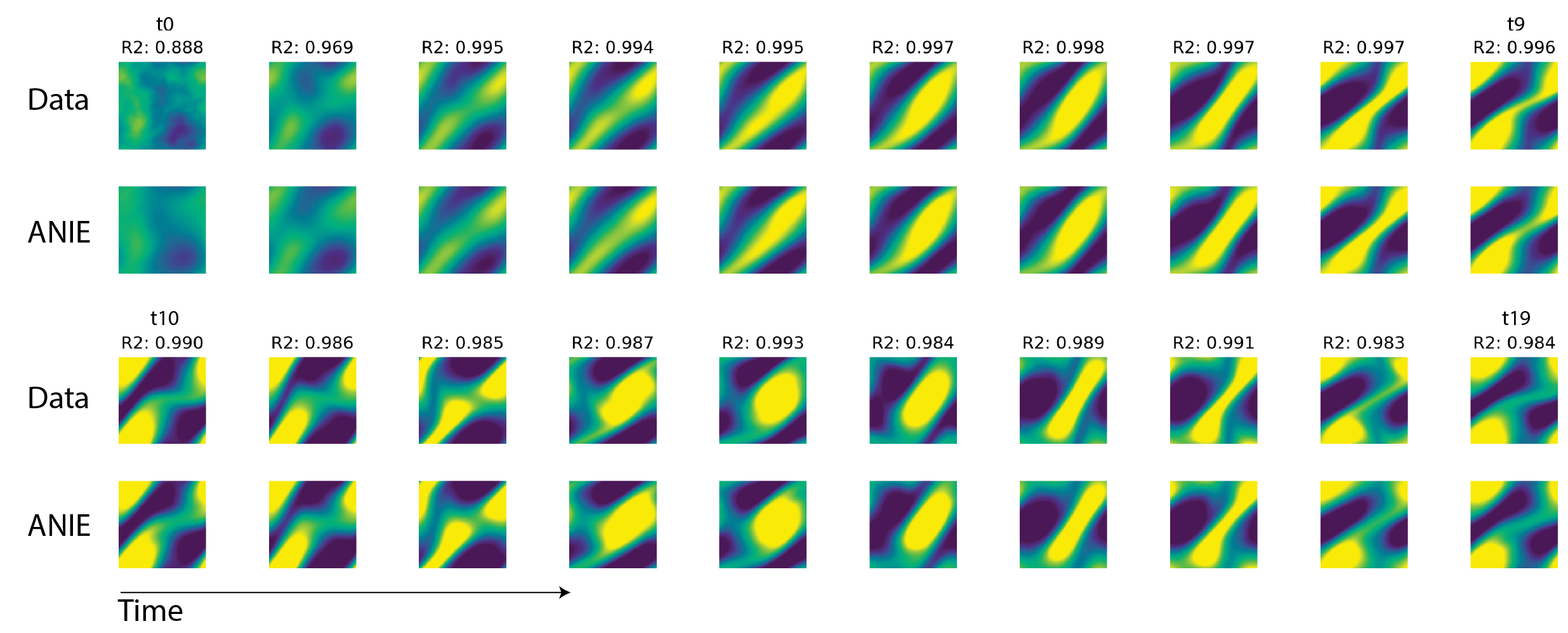}\caption{Example dynamics of Navier-Stokes system. Ground truth data (top) and prediction using ANIE (bottom) are shown. Prediction was generated using an initial condition that was not seen during training. R2 values quantify the model fit.}\label{fig:NS_output}
\end{figure*}

\begin{figure*}[h]
\centering
  \includegraphics[width=0.65\textwidth]{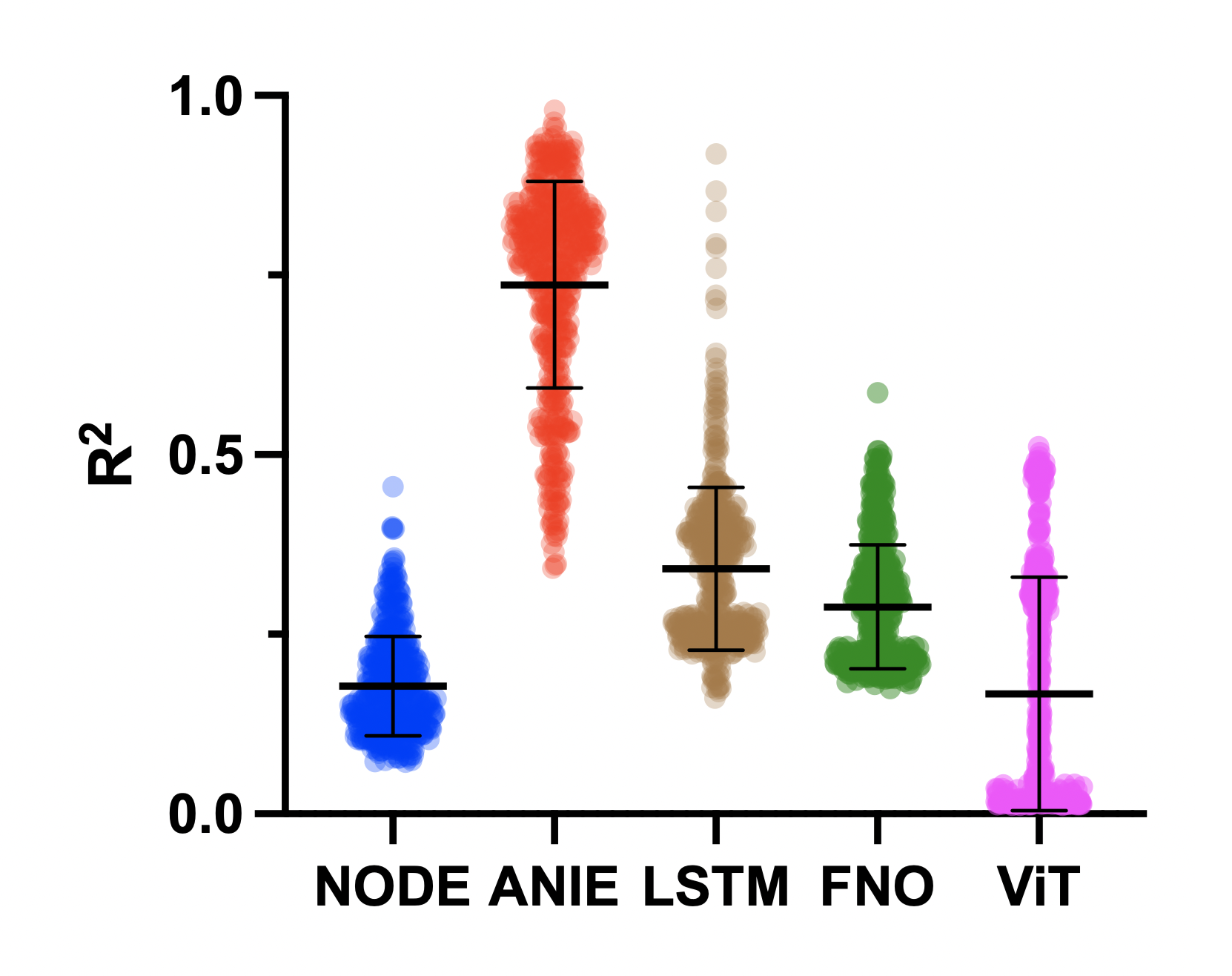}\caption{Quantification, using R-squared, of model fits to 2D IE spiral dataset. Models were run during inference on initial conditions not seen during training. ANIE has the best performance (highest R-squared) in predicting the dynamics.}\label{fig:CorrelationCoeff_2d_IE_spirals} 
\end{figure*}


\begin{figure*}[ht]
\centering
  \includegraphics[width=0.75\textwidth]{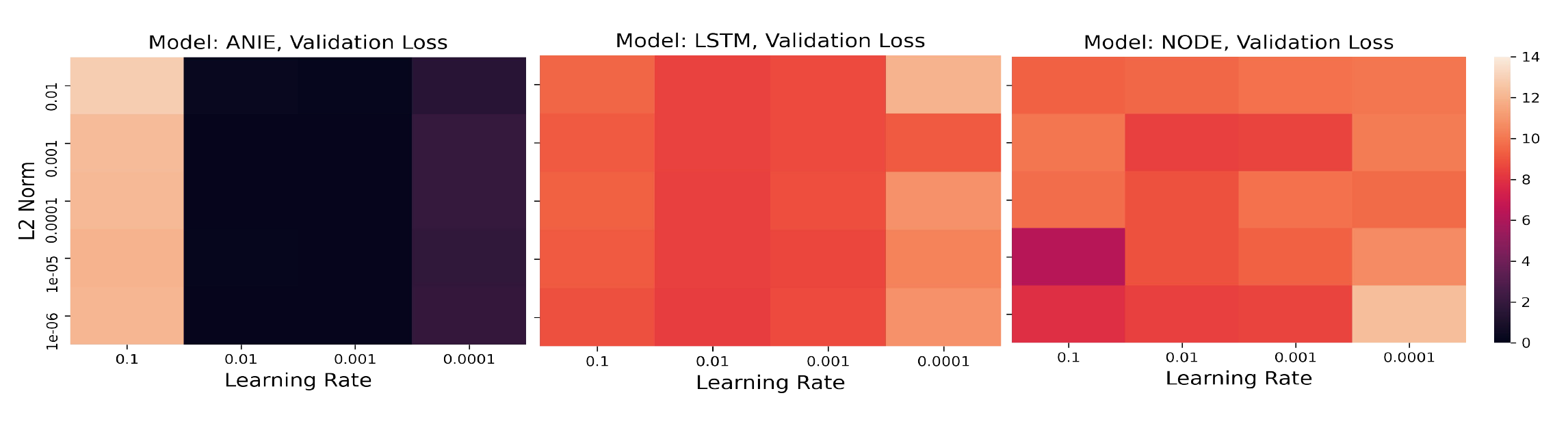}\caption{Hyperparameter sensitivity analysis for ANIE, LSTM and Neural ODE. Validation set Mean Squared Error for different hyperparameter combinations for all models trained for 300 epochs.}\label{fig:sensitivity} 
\end{figure*}

\begin{figure*}[h]
\centering
  \includegraphics[width=0.75\textwidth]{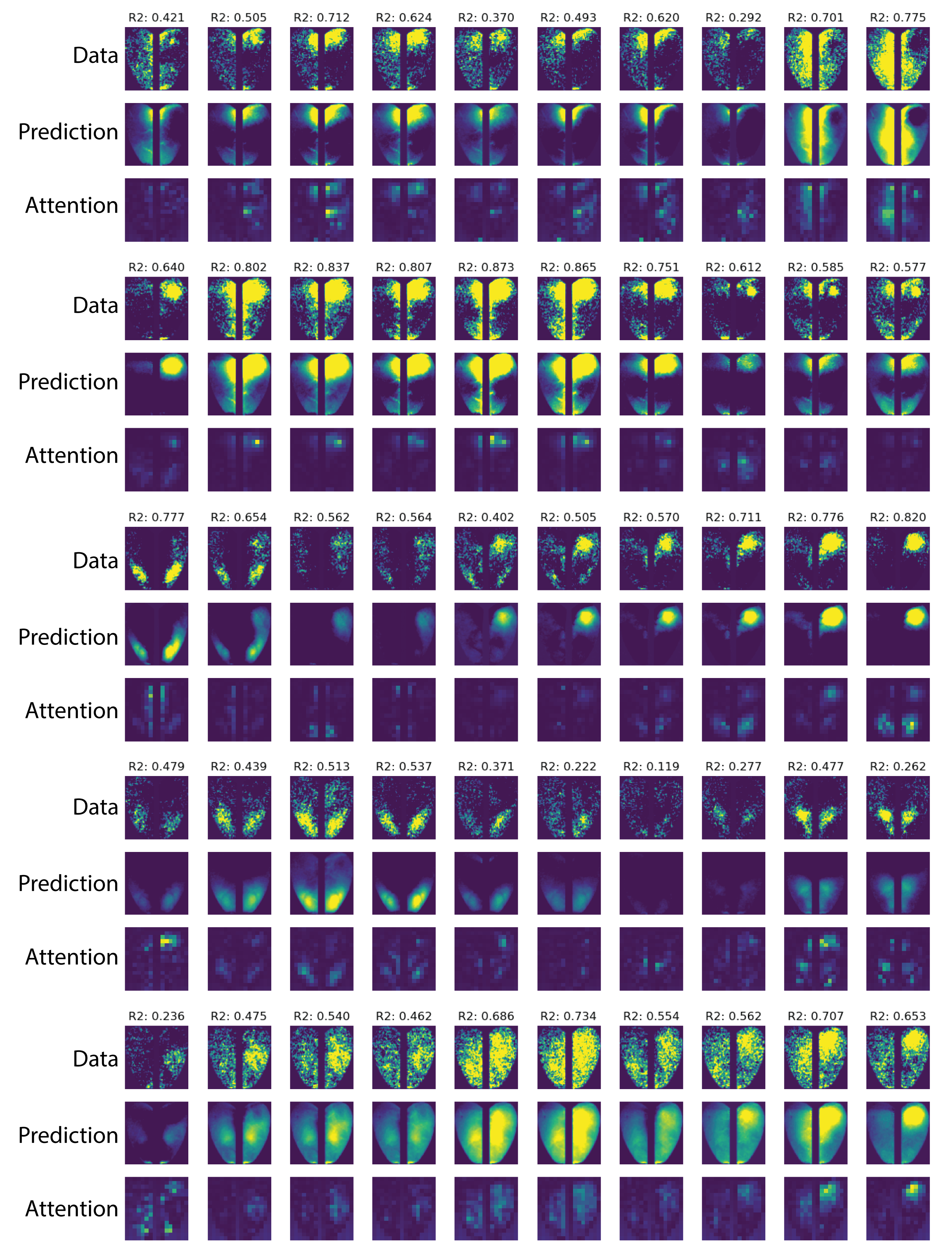}\caption{Example dynamics for the calcium imaging dataset, and their respective attention plots. We see that the attention weights do not directly reflect the input intensity and show activity for the motor and visual cortices.}\label{fig:ANIE_attention}
\end{figure*}

\begin{table}
\caption{Benchmark on embedding experiment. We perform KNN regression with $k = 5$ on embeddings of Navier-Stokes dynamics correlating the velocity of the dynamics and the embedding. All values are mean squared errors and are multiplied by a factor of $10^{-4}$.
}\label{tab:embedding_knn} 
\centering
\begin{tabular}{|c | c | c | c | c | }
  \hline 
  PCA& UMAP & LSTM & ViT  &\textbf{ANIE (ours)}\\
  \hline 
   $9.91\pm 2.30$ & $7.27 \pm 1.75$ & $8.83 \pm 1.55 $& $11.04 \pm 2.35$ & $\mathbf{6.52 \pm 1.31}$\\ 
   \hline
\end{tabular}\\
\end{table}\

\begin{table}[t]
\caption{Performance in $R^2$ of a KNN Regressor in regressing the contrast of visual stimuli from the learned latent representation. Results presented as (mean $\pm$ std, N=1600 frames, cross-validation=10).}
\label{tab:regression_visual_stim}
\begin{center}
\begin{tabular}{|c|c|c|c|c|}
\hline
  PCA & ViT & ConvLSTM & \textbf{ANIE (ours)} \\
\hline
$0.6763 \pm 0.02100$ & $0.6231 \pm 0.07581$ & $0.6263 \pm 0.06008$ & $\mathbf{0.6944 \pm 0.02982}$\\
\hline
\end{tabular}
\end{center}
\vskip -0.1in
\end{table}

\subsection{Benchmark of (A)NIE training speed}\label{subsec:speed}

Neural ODEs (NODEs) can be slow and have poor scalability \cite{kelly2020learning}. As such, several methods have been introduced to improve their performance \cite{kelly2020learning,kidger2021hey,daulbaev2020interpolation, poli2020hypersolvers,pal2021opening}. Despite these improvements, NODE is still significantly slower than discrete methods such as LSTMs. We hypothesize that (A)NIE has significantly better scalability than NODE, comparable to fast but discrete LSTMs, despite being a continuous model. To test this, we compare NIE and ANIE to the latest optimized version of (latent) NODE \cite{rubanova2019latent} and to LSTM on three different dynamical systems: Lotka-Volterra equations, Lorenz system, and IE-generated 2D spirals (see Appendix~\ref{data_generation} for data generation details). During training, models were initialized with the first half of the data and were tasked to predict the second half. Training speed are reported in Table~\ref{tab:walltime}. While all models achieve comparable (good) fits to the data, we find that ANIE outperforms all models in 2 out of the 3 datasets in terms of speed. Furthermore, ANIE has better MSE compared to all other models.

\begin{table*}[h]\caption{Comparison of training speed between NIE, ANIE, NODE, and LSTM on three different dynamical systems. Wall Time is provided in seconds per training iteration. Mean Squared Error shows accuracy of model fits to the data. NIE, ANIE, and LSTM are significantly faster than NeuralODE for all systems. However, while LSTM is fast, it is not a continuous time model like NIE, ANIE, and NeuralODE. Thus, NIE and ANIE have the advantages of being true continuous models with the speed of an LSTM. 
}\label{tab:walltime}

\centering
\resizebox{\textwidth}{!}{\begin{tabular}{|c | c | c | c| c| c| c|}
 \hline
  & \multicolumn{3}{|c|}{Wall Time (sec per iteration)} & \multicolumn{3}{|c|}{Mean Squared Error} \\ 
  \hline
 Models & Lotka-Volterra & Lorenz & 2D-Spirals & Lotka-Volterra & Lorenz & 2D-Spirals \\
 \hline
 LSTM & $0.0116$  & $\mathbf{0.012}$  & $0.02$ & $0.4492$ & $0.9926$ & $0.1313$ \\
 \hline
 Latent NeuralODE  & $2.88$  & $0.845$  & $0.494$ & $0.4502$ & $1.1001$ & $0.1385$  \\
 \hline
 \textbf{NIE (ours)} & $0.0312$ & $0.38$  & $0.154$ & $0.4596$ & $1.0229$ & $0.1449$  \\
 \hline
 \textbf{ANIE (ours)} & $\mathbf{0.0081}$  & $0.1764$ & $\mathbf{0.0082}$ & $\mathbf{0.4411}$ & $\mathbf{0.9421}$  & $\mathbf{0.1257}$  \\
 \hline
\end{tabular}}\\
\end{table*}\

\subsection{Hyperparameter sensitivity benchmark}\label{subsec:sensitive}

For most deep learning models, including NODE, finding numerically stable solutions usually requires an extensive hyperparameter search. Since IE solvers are known to be more stable than ODE solvers, we hypothesize that (A)NIE is less sensitive to hyperparameter changes. To test this, we quantify model fit, for the Lotka-Volterra dynamical system, as a function of two different hyperparameters: learning rate and L2 norm weight regularization. We perform this experiment for three different models: LSTM, Latent NODE, and ANIE. As shown in Figure~\ref{fig:sensitivity}, we find that ANIE generally has lower validation error as well as more consistent errors across hyperparameter values, compared to LSTM and NODE, therefore validating our hypothesis.

\subsection{Modeling 2D IE spirals}\label{subsec:2d_IE}

To further test the ability of ANIE in modeling non-local systems, we benchmark ANIE, NODE, and LSTM on a dataset of 2D spirals generated by integral equations. This data consists of 500 2D curves of 100 time points each. The data was split in half for training and testing. During training, the first 20 points were given as initial condition and the models were tasked to predict the full 100 point dynamics. Details on the data generation are described in Appendix~\ref{data_generation}. For ANIE, the initialization is given via the free function $f$, which assumes the values of the first 20 points and sets the remaining 80 points to be equal to the value of the 20th point. For NODE, the initialization is given as the reverse RNN on the first 20 points, which outputs a distribution corresponding to the first time point (see \cite{NODE} for more details on their Latent ODE experiments). For LSTM, we input the data in segments of 20 points to predict the consecutive point of the sequence. The process is repeated with the output of the previous step until all points of the curve are predicted. During inference, we test the models' performance on never before seen initial conditions. Table \ref{tab:corr_coeff_IE_2D_spirals} shows the correlation between the ground truth curve and the model predictions. Figure \ref{fig:CorrelationCoeff_2d_IE_spirals} shows the correlation coefficients for the 500 curves. In summary, ANIE significantly outperforms the other tested methods in predicting IE generated non-local dynamics.

\begin{table}
\caption{Benchmark on $2D$ IE spirals. R2 values of model fits are provided for ANIE, NODE, and LSTM. ANIE has the best performance.
}\label{tab:corr_coeff_IE_2D_spirals} 
\centering

\begin{tabular}{|c | c | c | c | c | }
  \hline 
  NODE & LSTM & ViT & FNO  &\textbf{ANIE (ours)}\\
  \hline 
   $0.1778 \pm 0.06932$ & $0.3410 \pm 0.1132$ &$0.1668 \pm 0.1624 $& $0.2882 \pm 0.08609$ & $\mathbf{0.7366 \pm 0.1440}$\\ 
   \hline
\end{tabular}\\
\end{table}\

\subsection{Solver convergence}\label{sec:convergence}

We now consider the convergence of the solver to a solution of an IE for a trained model. Our experiment here considers a model that has been trained with a number of iteration, and we explore whether the solver iterations converge to a solution at the end of the training. These results show that the model learns to converge to a solution of Equation~\ref{eqn:IE_operator_form} within the iterations that are fixed during training. They show that a fixed point for IE is obtained when outputting a prediction. 

\begin{figure*}[ht]
\centering
  \includegraphics[width=0.8\textwidth]{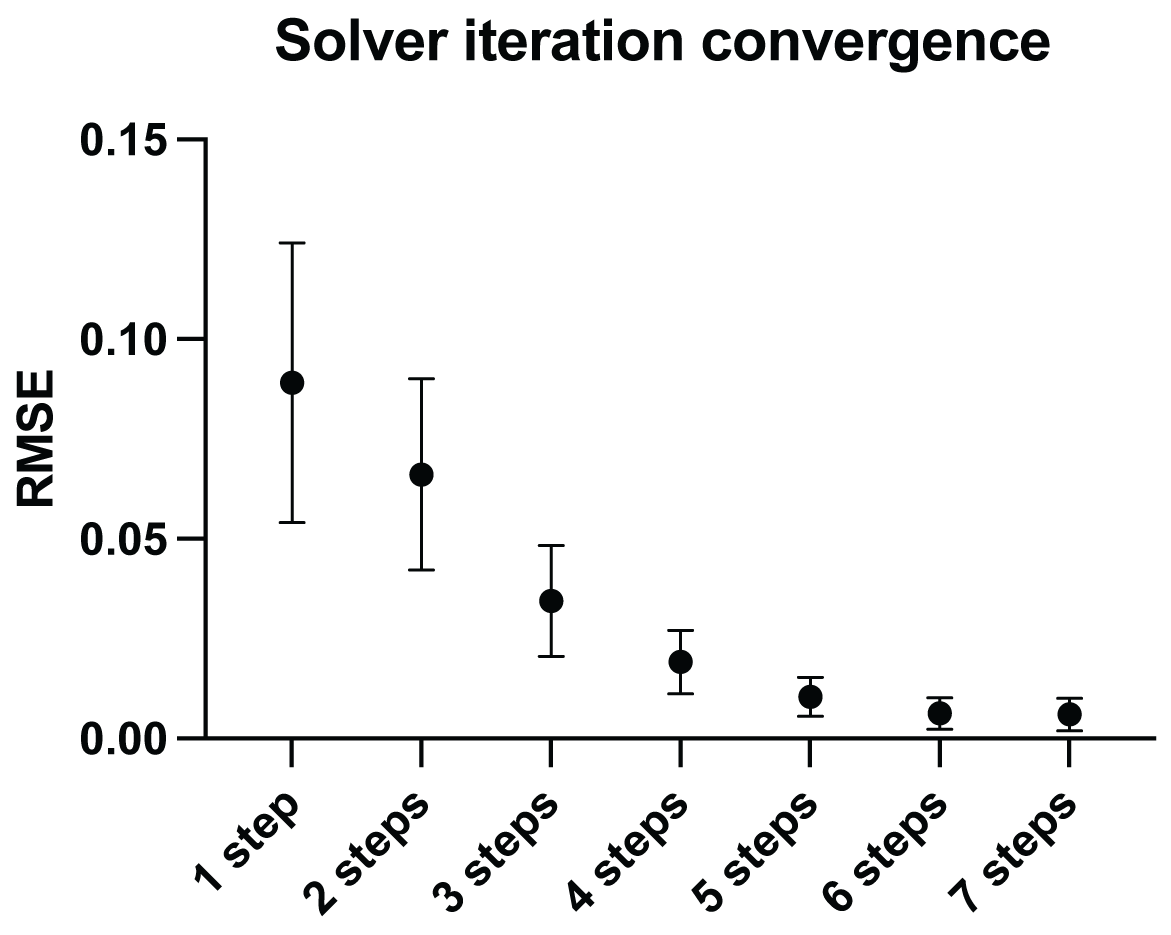}\caption{Rate of convergence during the iterations of the solver steps for a model trained on Navier-Stokes equations with $7$ iterations. The model learns to converge within the given steps.}\label{fig:iter_conv}
\end{figure*}


Figure~\ref{fig:iter_conv} and Figure~\ref{fig:iter_error} show the convergence error (i.e. the value $||y_{n+1} - y_n||$), and the guesses produced by the solver during inference (i.e. $y_n$ for $n$ corresponding to the iteration index), respectively.


\section{Integral Equations}\label{sec:Integral_general}

Integral Equations (IEs) are equations where the unknown function appears under the sign of integral. These equations can be given in general as 
\begin{equation}\label{eqn:IE_operator_form}
   \lambda \mathbf y = f + T(\mathbf y),
\end{equation}
where $T$ is an integral operator, e.g. as in Equation~\ref{eqn:gen_NIE} and Equation~\ref{eqn:PIE}, and $f$ is a known term of the equation. In fact, this functional equations have been studied for classes of compact operators $T$ that are not necessarily in the form of integral operators -- see \cite{brascamp1969fredholm} and references therein. We can distinguish two fundamental kinds of equations from the form given in Equation~\ref{eqn:IE_operator_form} that have been studied extensively throughout the years. When $\lambda = 0$, we say that the corresponding IE is of the {\bf first kind}, while when $\lambda \neq 0$ we say that it is of the {\bf second kind}. 

In this article we formulate our methods based on equations of the second kind for the following important theoretical considerations which apply to the case where $T$ is bounded over an infinite space (such as the space of functions as we consider in the article). Firstly, an equation of the first kind can easily have no solution, as the range of a bounded operator $T$ on an infinite space is not the whole space (see Proposition~4.41 in \cite{moretti2013spectral}). Therefore, for choices of $f$, there is no $\mathbf y$ such that $T(\mathbf y) = -f$ and therefore Equation~\ref{eqn:IE_operator_form} has no solutions. The other issue is intrinsic to the nature of equation of the first kind, and does not relate to the existence of solutions. In fact, any compact injective operator $T$ (on an infnite space) does not admit a bounded left inverse (see Proposition~4.42 in \cite{moretti2013spectral}). In practice, this means that if Equation~\ref{eqn:IE_operator_form} has a unique solution for $f$, then varying $f$ by a a small amount can result in very significant variations in the corresponding solution $\mathbf y$. This is clearly a potential issue when dealing with a deep learning model that aims at learning the operator $T$ from data. In fact, observations from which $T$ is learned might be noisy, which might result in very considerable perturbations of the solution $\mathbf y$ and, consequently, considerable perturbations on the operator $T$ that the model converges to. Since equations of the second kind are much more stable, we have formulated all the theory in this setting, and implemented our solver for such equations. The issues relating to existence and uniqueness of the solution for these equations are discussed in more detail in Section~\ref{sec:existence} below.

The theories of IEs and Integro-Differential Equations (IDEs) are tightly related, and it is often the case to reduce problems in IEs to problems in IDEs and vice versa, both in practical and theoretical situations. IEs are also related to differential equations, and it is possible to reformulate problems in ODEs in the language of IEs or IDEs. In certain cases, IEs can also be converted to differential equations problems, even though this is not always possible \cite{Zem,GIKM}. In fact, the theory of IEs is not equivalent to that of differential equations. The most intuitive way of understanding this is by considering the local nature of differential euqations, as opposed to the non-local origin of IEs. By non-locality of IEs it is meant that each spatiotemporal point in an IE depends on an integration over the full domain of the solution function $\bold y$. In the case of differential equations, each local point depends only on the contiguous points through the local definition of the differential operators.

\subsection{Integral Equations (1D)}

We first discuss IEs where the integral operator only involves a temporal integration (i.e. 1D), as discussed in Section~\ref{sec:NIE}. In analogy with the case of differential equations, this case can be considered as the one corresponding to ODEs. 

These IEs are given by an equation of type 
\begin{equation}\label{eqn:gen_NIE_appendix}
    \bold y(t) = f(t) + \int_{\alpha(t)}^{\beta(t)} G(\bold y,t,s) ds,
\end{equation}
where $f$ is the free term, which does not depend on $\bold y$, while the unknown function $\bold y$ appears both in the LHS, and in the RHS under the sign of integral. The term $\int_{\alpha(t)}^{\beta(t)} G(\bold y,t,s) ds$ is an integral operator $\mathcal C(D) \longrightarrow \mathcal C(D)$ from the space of integrable functions $\mathcal C(D)$ over some domain of $\mathbb R$, into itself. We observe that the variables $t$ and $s$ appearing in $G$ are both in $D$, and they are interpreted as time variables. We refer to them as {\it global} and {\it local} times, respectively, following the same convention of \cite{NIDE}. The functions $\alpha$ and $\beta$ determine the extremes of integration for each (global) time $t$. Common choices for $\alpha$ and $\beta$ include $\alpha(t) = 0$ and $\beta(t) =t$ (Volterra equations), or $\alpha(t) = a$ and $\beta(t) = b$ (Fredholm equations). 

The fundamental question in the theory of IEs is whether solutions exists and are unique. It turns out that under relatively mild assumptions on the regularity of $G$ IEs admit unique solutions \cite{Zem}. Furthermore, the proofs in the first chapter of \cite{Lak} show the close relation between IEs and IDEs, as existence an uniqueness problems for IDEs are shown to be equivalent to analogous problems for IEs. Then the fixed point theorems of Schauder and Tychonoff are used to prove the results.

\subsection{Integral Equations (n+1D)}

We now discuss the case of IEs where the integral operator involves integration over a multi-dimensional domain of $\mathbb R^n$. This is the IE version of PDEs, and they are commonly referred to as Partial Integral Equations (PIEs) when integration occurs on different components separately. We will consider equations where the multi-dimensional integral is obtained through multiple integrations. An equation of this type takes the form 
\begin{equation}\label{eqn:PIE_appendix}
    \bold y(\bold x,t) = f(\bold x,t) + \int_{\Omega}\int_{\alpha(t)}^{\beta(t)} G(\bold y, \bold x, \bold x', t, s) d\bold x' ds,
\end{equation}
where $\Omega \subset \mathbb R^n$ is a domain in $\mathbb R^n$, and $\bold y: \Omega \times \mathbb R\longrightarrow \mathbb R^m$. Here $m$ does not necessarily coincide with $n$. 

PIEs and higher dimensional IEs have been studied in some restricted form since the 1800's, as they have been employed to formulate the laws of electromagnetism before the unified version of Maxwell's equations was published. In addition, early work on the Dirichelet's problem found the IE approach proficuous, and it is well known that several problems in scattering theory (molecular, atomic and nuclear) are formulated in terms of (P)IEs. In fact, the Schr\"{o}dinger equation can be recast as an IE \cite{tobocman1959integral}. See also \cite{salpeter1951relativistic}, where bound-state problems are treated with the IE formalism. 

\subsection{Generalities on solving Integral Equations}\label{sec:solving}

The most striking difference between the procedure of solving an IE and an ODE, is that for an IE in order to evaluate at a single time point, one needs to know the solution for all time points. This is clearly an issue, since solving for one point requires that we already know a solution for all points. In order to better elucidate this point, we consider a simple comparison between the solution procedure of an ODE equation of type $\dot{\mathbf y} = f(\mathbf y,t)$, and an IE of type $\mathbf y = f(t) + \int_0^1G(\mathbf y,t,s)ds$. 

Let us assume we are solving an ODE of type $\dot{\mathbf y}(t) = f(\mathbf y,t)$ and that $\mathbf y$ is known at time points $t_0, t_1, \ldots, t_{k-1}$. Then one can obtain $\mathbf y$ at $t_k$ by means of the Euler method by using the known value at $t_{k-1}$ by taking small enough steps $\Delta t$ forward in time. In general, therefore, one starts by the initial condition $\mathbf y_0$ and determines the solution $\mathbf y$ at the points $t_0, \ldots , t_n$ by taking small steps and representing the derivative as $\Delta f/\Delta t$ for small intervals $\Delta t$. Of course, more sophisticated methods are possible for the numerical solution of the ODE, but they essentially produce the next time point from the previous one in a sequential way. Let us now consider an analogous Fredholm IE to the ODE given above. This is a simple equation of type $\mathbf y = f(t) + \int_0^1G(\mathbf y,t,s)ds$. Suppose we know $\mathbf y$ at time points $t_0, \ldots , t_{k-1}$. In order to determine $\mathbf y(t_k)$, we need to compute $f(t_k)+\int_0^1G(\mathbf y,t_k,s)ds$, which requires us to know $\mathbf y$ over the full interval $[0,1]$, as $G$ is integrated over $[0,1]$. It is obvious that knowing a single time point for $\mathbf y$ (or a sequence of values) does not suffice anymore. In a Volterra type of equation the integral would be between $[0,t_k]$ (where the unknown value $t_k$ is included!), which does not really change the essence of the issue. 

Although several methods can be employed to solve IEs, most (if not all) of them are based on the concept of iteration over some initial guess for the solution of the IE. Iterating on the initial guess produces a sequence of functions that then converges to a solution of the integral equation. More specifically, one can consider the von Neumann series of the integral operator as we now discuss. In fact, let us consider Equation~\ref{eqn:IE_operator_form}, which can be rewritten as
$$
(\mathbb 1 - T)(\mathbf y) = f,
$$
where we assume for a moment that $T$ is a linear operator.
Observe that if we can find the inverse of $\mathbb 1 - T$, then we obtain $\mathbf y$ as $(\mathbb 1 - T)^{-1}(f)$. This can be done by writing the von Neumann series for $(\mathbb 1 - T)^{-1} = \sum_{k=0}^{\infty} T^k$. This expression makes sense whenever the series $\sum_{k=0}^{\infty} T^k$ converges in operator norm, which is guaranteed in important cases such as when $\sum_{k=0}^{\infty} ||T||^k$ converges (e.g. when $||T||<1$), while milder conditions on the convergence of the series exist as well. In such a situation when the von Neumann series is meaningful, we can then obtain $\mathbf y$ by iteratively applying $T^k$ to $f$. The nonlinear case is handled in a similar iterative procedure, which is called {\it Method of Successive Approximations}, or also {\it Picard's Iterations} \cite{davis1960introduction}. It is in fact straightforward to convince oneself that under mild conditions, the method will output a solution of the integral equation. Conditions under which such succession is guaranteed to converge can be found in \cite{davis1960introduction}. A particularly well known case is when the integrand of the integral operator is contractive (i.e. Lipschitz with constant between $0$ and $1$) with respect to the variable $\mathbf y$. We give a proof of such approach for our setting, c.f. similar results found in \cite{davis1960introduction}. 

\begin{theorem}\label{thm:iteration_method}
    Let $\epsilon>0$ be fixed, and suppose that
    $T$ is Lipschitz with constant $k<1$. Then, we can find $y\in X$ such that $||T(y) + f - y|| < \epsilon$, independently of the choice of $f$. 
\end{theorem}
\begin{proof}
     Let us set $y_0 := f$ and $y_{n+1} = f + T(y_n)$ and consider the term
     $||y_1 - y_0||$. We have
     $$
     ||y_1-y_0|| = ||T(y_0)|| =  ||T(y_0)||.
     $$
     For an arbitrary $n>1$ we have
     $$
     ||y_{n+1} - y_n|| = ||T(y_n) -  T(y_{n-1})|| \leq k ||y_n - y_{n-1}||.
     $$
     Therefore, applying the same procedure to $y_n-y_{n-1} = T(y_{n-1}) - T(y_{n-2})$ until we reach $y_1 - y_0$, we obtain the inequality
     $$
     ||y_{n+1} - y_n|| \leq k^n ||T(y_0)||. 
     $$
     Since $k <1$, the term $ k^n ||T(y_0)||$ is eventually smaller than $\epsilon$, for all $n \geq \nu$ for some choice of $\nu$. Defining $y:= y_\nu$ for such $\nu$ gives the following
     $$
     ||T(y_\nu) + f - y_{\nu} || = ||y_{\nu+1} - y_\nu|| < \epsilon. 
     $$
\end{proof}

The following now follows easily.
\begin{corollary}\label{cor:iterative_method}
    Consider the same hypotheses as above. Then Equation~\ref{eqn:IE_operator_form} admits a solution. In particular, if the integrand $G$ in Equation~\ref{eqn:gen_NIE_appendix} is contractive with respect to $\mathbf y$ with constant $k$ such that $k\cdot (b-a) < 1$ (where $[a,b]$ is the codomain of $\alpha, \beta$), the iterative method in Algorithm~\ref{algo:NIE} converges to a solution of the equation.
\end{corollary}
\begin{proof}
    From the proof of Theorem~\ref{thm:iteration_method} it follows that the sequence $y_n$ is a Cauchy sequence. Since $X$ is Banach, then $y_n$ converges to $y\in X$. By continuity of $T$, $y$ is a solution to Equation~\ref{eqn:IE_operator_form}. For the second part of the statement, observe that when $G$ is contractive with respect to $\mathbf y$, then we can apply Theorem~\ref{thm:iteration_method} to show that the sequence generated following Algorithm~\ref{algo:NIE} is Cauchy, and we can proceed as in the first part of the proof. 
\end{proof}

\begin{remark}\label{rmk:iterative_method}
{\rm 
Observe that the result in Corollary~\ref{cor:iterative_method} applies to Algorithm~\ref{algo:ANIE} as well, under the assumptions that the transformer architecture is contractive with respect to the input sequence $\mathbf y$. Also, a statement that refers to higher dimensional IEs can be obtained (and proved) similarly to the second part of the statement of Corollary~\ref{cor:iterative_method}, using the measure of $\Omega\times [a,b]$ instead of the value $(b-a)$. 
}
\end{remark}

In practice, the method of successive approximations is implemented as follows.
The initial guess for the integral equation is simply given by the free function $f$ (i.e. $T^0(f)$), which is used to initialize the iterative procedure. Then, we apply $T$ to $\mathbf y^0 := T^0(f)$ to obtain a new solution $\mathbf z^1 := f(t) + T^1(\mathbf y^0)$. We set $y^1 := r\mathbf y^0 + (1-r)\mathbf z^1$ and apply $T^2$ to the solution $y^1$ and repeat. Here $r$ is a smoothing factor that determines the amount of contribution from the new approximation to consider at each step. As the iterations grow, the fractions of the contributions due to the smoothing factor $r$ tend to $1$. Observe that when we sum $r\mathbf y^i + (1-r)\mathbf y^{i+1}$ with $r=0$, we obtain the terms of the von Neumann series up to degree $i$ applied to $f$: $\sum_0^i T^k(f)$. The smoothing factor generally shows good empirical regularization for IE solvers, and we have set $r=1/2$ throughout our experiments, even though we have not seen any concrete difference between different values of $r$. This procedure is shown in Fugure~\ref{fig:Solver}.


In \cite{delves1988computational}, Chapter 4, computations on the error bounds for the iterative procedure described above when the integrand function $G$ splits into the product of a kernel (see above) and a linear function $F$ are given. In the same chapter a detailed description of the Nystr\"{o}m approximation for the computation of the error is given as well. We describe a concrete realization of the iterative procedure discussed above in Section~\ref{sec:solver}, along with the learning steps for the training of our model. Moreover, we additionally observe that the procedure described above does not depend on $T$ being an integral operator or a general operator and, therefore, applying this methodology to the case where we have a transformer instead of $T$ is still meaningful, in the assumption that $T$ is such that the iterated series of approximations is convergent. 

Depending on the specific IE that one is solving (e.g. Fredholm or Volterra, $1D$ or $(n+1)D$ etc.) the actual numerical procedure for finding a numerical solution can vary. For instance, a list of articles that showcase such a wide variety of specific methods for the solution of certain types of equations is \cite{kushnir2012highly,borowko2017integral,li2002solution,kazemi2019new,parand2020solving}. Such variations upon the same theme of iterative procedure depend on finding the most efficient way of converging to a solution, finding the best error bounds, improving stability of the solver and substantially depend on the form of the integral operator. As our method is applied without the actual knowledge of the shape of the integral operator, but it actually aims at inferring (i.e. learning) the integral operator from data, we implement an iterative procedure which is fixed and depends only on a hyperparameter smoothing factor. This is described in detail in the next section. However, we point out that since the integrand, and therefore the integral operator itself, is learned during the training, one can assume that the model will optimize with respect to the procedure in a way that our iterations are in a sense ``optimal'' with respect to the target. 

Thus far, our considerations on the implementation of IE solvers seem to point to a fundamental computational issue, since they entail a more sophisticated solving procedure than that of ODEs or PDEs. However, in various situations, even solving ODEs and PDEs through IE solvers presents significant advantages that are not necessarily obvious from the above discussions. The first advantage is that IE solvers are significantly more stable than ODE and PDE solvers, as shown for instance in the articles \cite{kushnir2012highly,rokhlin1985rapid,rokhlin1990rapid}. This in particular provides a new solution to the issue of underflowing during training of NODEs that does not consist of a regularization, but of a complete change of perspective. In addition, even though to solve an IE one needs to iterate, in general the number of iterations is not particularly high. In fact, in our experiments the total number of iterations turned out to be sufficient to be fixed between $4$ and $6$. However, when solving for instance an ODE, one needs to go sequentially through each time step. These can be in the order of the $100$ (as in some of our experiments). On the contrary, our IE solver processes the full time interval in parallel for each iteration. This results in a much faster algorithm compared to differential solvers, as shown in our experiments.

\subsection{Existence and uniqueness of solutions}\label{sec:existence}

The solver procedure described in the previous subsection of course assumes that there exists a solution to start with. As mentioned at the beginning of the section, we treat equations of the second kind in this article also because the existence conditions are better behaved than for the equations of the first kind. We give now some theoretical considerations in this regard. We will also discuss when these solutions are uniquely determined. Existence and uniqueness are two fundamental parts of the well-posedness of IEs, the other being the continuity of solutions with respect to initial data.

A concise and relatively self-contained reference for the existence and uniqueness of solutions for (linear) Fredholm IEs is Section~4.5 in \cite{moretti2013spectral}. In fact, it is shown in Theorem~4.43 that if a Fredholm equation has Hermitian kernel, then the IE has a unique solution whenever $\lambda$ is not an eigenvalue of the integral operator. For real coefficients, which is the case we are interested into, one can simply reduce the case to symmetric kernels, which are kernels for which $K(t,s) = K(s,t)$ for all $t,s$. Since in this article we have assumed $\lambda = 1$, the condition becomes equivalent to saying that there is no function $\mathbf z$ such that $\int_0^1K(t,s)z(s)ds = z(t)$ for all $t$. 

For more general (linear) integral operators (bounded on a Hilbert space), a similar result holds. In fact, from Theorem~4.44 in \cite{moretti2013spectral} we have that a generalized Fredholm integral equation admits solutions if and only if the free function is orthogonal to each solution of the associated homogeneous adjoint equation. The latter admits the zero function as a solution (so the solution set is not empty), and is obtained from Equation~\ref{eqn:IE_operator_form} by deleting $f$, and by taking the adjoint of $T$ and the complex conjugate of $\mathbf y$. In the real case, the conjugate of $\mathbf y$ is $\mathbf y$ itself. Moreover, uniqueness is guaranteed if the associated homogeneous equation has only trivial solutions. In the case of nonlinear integral operators several existence and uniqueness conditions can be found in the literature. The reader is referred to \cite{davis1960introduction,Zem,Lak} for specific formulations. Generally speaking, such conditions are assumed on the integrand functions that determine the integral operator, in such a way that contractive theorems (such as Schauder's and Tikhonoff's) can be applied. 

Observe that such formulations of the existence and uniqueness based on the contractive properties of the operator $T$ are particularly interesting in the case where the integral operator is replaced by a general neural network (between function spaces) which is not necessarily obtained through integration. In practice, when $T$ is a general neural network that is possibly nonlinear on all the entries, except with respect to the function $\mathbf y$, $T$ can be approximated by an integral equation using the following reasoning.
It is known that Hilber-Schmidt operators on the Hilbert space of square integrable functions are approximated by integral operators -- see e.g. \cite{moretti2013spectral} Theorem 4.26. It is reasonable to assume that neural network operators are well behaved enough to be considered Hilbert-Schmidt operators. They therefore approximate some integral operator, and the training process therefore learns an integral equation.

More generally, for IEs of Urysohn or Hammerstein type, the existence and uniqueness problems are well known under much milder conditions, namely when the operator is completely continuous \cite{Topological,Geometrical}. In this situation, it is sufficient for the operator to have a nonzero topological index to guarantee that the corresponding IE admits a solution, and to study the problem of uniqueness one can determine the value of the topological index in a bounded subset of the Banach space in consideration, since this is directly related to the number of fixed points of the given IE. 

The previous discussion, however, does not directly apply to the case when $T$ is a transformer, due to the fact that $T(\mathbf y)$ is not linear in $\mathbf y$ in this case. Such equations can still be considered generalized Fredholm equations, and conditions on nonlinear operators $T$ being approximated by integral operators can be found in the literature, but the extent to which such equations are equivalent to integral equations is a fascinating question which will not be explicitly considered in this article. 

As a word of warning, we mention that the general theory ensures the existence and uniqueness of solutions under some (mild) assumptions. Of course, in principle one should impose constraints to ensure that such assumptions are satisfied and that the results would apply. However, in our experiments we have observed good stability and good convergence without imposing any additional constraints. This does not apply in general, but we hypothesize that during optimization the model converges towards operators whose associated IE is well behaved, in order to avoid regimes of poor stability due to lack of solutions or lack of uniqueness of solutions. For different datasets such behavior might not be satisfied, and extra care in this regard might be needed.

\subsection{The initial condition for IEs}

NIE does not learn a dynamical system via the derivative of a function $\bold y$, as is the case for ODEs and IDEs. Therefore, we do not need to specify an initial condition in the solver during training and evaluation. In fact, the initial condition for IEs is encoded in the equation itself. For instance, taking $t=0$ in a Volterra or a Fredholm equation uniquely fixes $\bold y(\bold x, 0)$ for all $x$.

Therefore, we can specify a condition for IEs by constraining the free function $f(\bold y,t)$. 
In what follows, we will make use of this paradigm several times. There are two immediate ways one could impose constraints on the free function. The simplest is to fix a value $\bold y_0$ and let $f(\bold y,t)$ be fixed to be $\bold y_0$ for all $t$. Alternatively, one could choose an arbitrary function $f$ and keep this function fixed. In practice, the latter is conceptually more meaningful. For instance, in theoretical physics, when transforming the Schr\"{o}dinger equation into an integral equation, on the RHS one can choose an arbitrary function $\psi(\bold y,t)$ which corresponds to the  wave function of free particles, i.e. without potential $V$. Applications of this procedure are found below in the experiments.

\section{Approximation capabilities}

In this appendix we consider the capabilities of our models to approximate (nonlinear) integral operators and integral equations.

\subsection{NIE}
We consider two settings, where the integral operarator is modeled by a single hidden layer feed forward neural network of arbitrary width, or by an arbitrarily deep neural network. 

We want to show that when we restrict ourselves to single hidden layer feed forward neural networks of arbitrary depth for our function $G_{\theta}$ in Equation~\ref{eqn:gen_NIE}, we can approximate a wide class of integral equations over a suitable subset of the space of functions. In the case of deep neural networks, we will argue that the NIE architecture can approximate any  ``regular enough'' integral operator, where regularity will be described below. We restrict our considerations to the case of function spaces where the domain is $\mathbb R$, since the higher dimensional case is easily adapted from this discussion. We will therefore use $y$ instead of $\mathbf y$ to indicate the elements of the domain of the integral operators.

Let $T: C([0,1]) \longrightarrow C([0,1])$ be an integral operator on the space of continuous functions, defined as $y \mapsto T(y)(t) := \int_{\alpha(t)}^{\beta(t)}G(y(s),t,s)ds$ for continuous functions $\alpha, \beta : [0,1] \longrightarrow [0,1]$, and a continuous $G: \mathbb R\times [0,1]\times [0,1]\longrightarrow \mathbb R$. In fact, for what follows we could consider $G$ as being Borel measurable, instead of imposing the more restrictive condition of being continuous. However, since in applications continuity is generally required, we impose this more restrictive conditions. Moreover, our discussion easily extends to the case when the definition intervals are $[a,b]$ instead of $[0,1]$ with simple modifications, and a similar approach also extends to higher dimensional integrals. We assume that $T$ is such that the corresponding integral equation of the second kind, i.e. Equation~\ref{eqn:gen_NIE}, admits a solution $y^*: [0,1] \longrightarrow \mathbb R$ in $C([0,1])$. Since $y^*$ is continuous, there exists a compact $K = [-k,k]$, for $k>0$, such that $y^*([0,1]) \subset K$. Let us consider now a neighborhood $U_K$ of $y^*$ in the compact-open topology such that for all $y\in U_K$ we have the property $y([0,1]) \subset K$. This could be, for instance, the space of functions $y$ mapping $[0,1]$ into the open $(-k,k) = K^\circ$.  We can therefore restrict $G$ to the domain $K\times [0,1]\times [0,1]$, and we will still indicate this restriction by $G$ and the corresponding integral operator by $T$ (defined over the neighborhood $U_K$), for notational simplicity. 

For an arbirtary chosen $\epsilon >0$, we want to show that we can approximate $T(y)$ with error at most $\epsilon$ in the metric induced by $C([0,1])$ on $U_K$ through an NIE integral operator $T_\theta(y)(t) := \int_{\alpha(t)}^{\beta(t)}G_\theta(y(s),t,s)ds$. To this purpose, let us set $Q = \sup_{[0,1]} |\beta(t) - \alpha(t)|$, and observe that by applying the Universal Approximation Theorem for single hidden layer feed forward neural networks (see \cite{hornik1989multilayer}) to the function $G: K\times [0,1] \times [0,1] \longrightarrow \mathbb R$, we can find a single hidden layer neural network $G_\theta: K\times [0,1] \times [0,1] \longrightarrow \mathbb R$ such that for all $t,s\in [0,1]$ we have $|G(y(s),t,s) - G_\theta(y(s),t,s)| < \epsilon/Q$. With such a $G_\theta$, for all functions $y\in U_K$ we have for any fixed $t^*$ in $[0,1]$ 
\begin{eqnarray*}
  || T(y)(t^*) - \int_{\alpha(t^*)}^{\beta(t^*)} G_\theta(y(s),t^*,s)ds ||  &\leq& \int_{\alpha(t^*)}^{\beta(t^*)}||G(y(s),t^*,s) - G_\theta(y(s),t^*,s)||ds \\ &<& |\beta(t^*) - \alpha(t^*)| \epsilon/Q. 
\end{eqnarray*}
Therefore, uniformly on $t$ we have that 
$$
|| T(y)(t) - \int_{\alpha(t)}^{\beta(t)} G_\theta(y(s),t,s)ds || < \epsilon. 
$$
But this means that $d(T(y),T_\theta(y))<\epsilon$ with the metric $d$ on $U_K$ induced by that of $C([0,1])$. 

We observe that while this approximation does not hold in complete generality, it is valid for a class of integral operators of importance, since we are usually interested in operators whose corresponding IE is admits continuous solutions, and we are interested in modeling the operator in a neighborhood of a solution. Moreover, under mild assumptions (e.g. discussed in Appendix~\ref{sec:existence} the dependence of the solution on the initial data is continuous, and therefore the solutions to the equation for perturbed $f$ lie in a neighborhood of a solution $y^*$ obtained for $f$. So, our results apply in such important cases. Lastly, we point out that throughout the previous reasoning we have implicitly assumed that numerical integration is performed with infinite precision. Of course this is not the case in practice, but since we can reduce the numerical error in the integration procedure arbitrarily by choosing densely enough samples for a choice of the integration scheme, the error due to numerical integration can be rendered small enough so that the previous inequalities hold.  

We now consider the case where we allow deep neural networks as in \cite{lu2017expressive}, Theorem~1. In this case, we argue that for any IE of the second kind as in Equation~\ref{eqn:gen_NIE} where we set $T(y)(t) := \int_{\alpha(t)}^{\beta(t)}G(y(s),t,s)ds$ for a Lebesgue integral function $G$, we can approximate the integral operator $T$ with arbitrary precision. As a consequence, there is a NIE model which realizes any IE as in Equation~\ref{eqn:gen_NIE} with arbitrary accuracy. We can proceed as for the case of single hidden layers neural networks above, with the main difference that applying Theorem~1 from \cite{lu2017expressive} we do not need to restrict ourselves to a neighborhood $U_K$ of a solution $y^*$ of the integral equation, and the neural integral operator $\int_{\alpha(t)}^{\beta(t)}G_\theta(y(s),t,s)ds$ approximates $T$ uniformly with respect to $t$ for any $y\in C([0,1])$. Observe that in order to use Theorem~1 in \cite{lu2017expressive} we need to precompose $G$ and $G_\theta$ by a characteristic function $\chi_{[0,1]}$, which does not affect the result. 

\subsection{ANIE}

We give some comments on the approximation properties of ANIE with respect to generalized Fredholm equations. For simplicity, we consider the case where the integration is performed only over time, even though the same reasoning can be extended to spatiotemporal domains. Let $T: C([0,1]) \longrightarrow C([0,1])$ denote a Fredholm integral operator defined through the assignment $T(y)(t) = \int_0^1 G(\mathbf y(t), t, \mathbf y(s), s) ds$. Observe that this integral form is more general than considered in Equation~\ref{eqn:gen_NIE}, and it follows the interpretation of integration in terms of self-attention (c.f. Appendix~\ref{sec:ANIE_solving_training}, where the integration approximation used in this article is given in more detail). 

Let us assume that the integral equation $y = f^* + T(y)$ admits a unique continuous solution $y^*\in C^2([0,1])$, and that $G$ is regular enough so that the equation admits unique solution in $C([0,1])$ for given functions $f$ in a neighborhood of $f^*$ in the compact open topology. Observe, that such well-posedness conditions are usually relatively mild (see for instance Appendix~\ref{sec:existence} and references therein), and this is the main situation of interest in applications. Then, there exists a compact $K = [-k,k]$ such that $y^*([0,1])\subset K$ and we can choose a neighborhood $U_K$ of $y^*$ in the compact-open topology of $C([0,1])$ such that $y([0,1]) \subset K$ for all $y\in U_K$. In fact, one can simply choose $U_K := \{y\in C([0,1])\ | \ |y([0,1])| < k \}$. 
Under such hypothesis, there are numerical integration schemes that can approximate the integral $\int_0^1 G(\mathbf y(t), t, \mathbf y(s), s) ds$ for any fixed choice of $t$ with arbitrary precision, upon choosing a number of points for evaluation that is sufficiently large. For instance, for any fixed $t$, the error for trapezoidal rules is bound by a term that goes to zero as $n$ grows, where $n$ is the number of points chosen in $[0,1]$ for approximating the integral, see Equation 2.1.6 in \cite{davis2007methods}. This term is the modulus of continuity 
$$\omega_t(1/n) := \max_{|s_1-s_2|<1/n} |G(\mathbf y(t), t, \mathbf y(s_1), s_1) - G(\mathbf y(t), t, \mathbf y(s_2), s_2)|.$$
For each choice of $n$, there exists a compact $K_n := [-k_n,k_n]$ such that $y^*$ maps into $K_n$, and $G$ restricted to $K_n\times [0,1]\times K_n\times [0,1]$ has $\omega_t(1/n) < 1/n$ for all $t\in K_n$. In this situation we can choose a neighborhood of $y^*$, $U_{K_n}$ such that $\omega_t(1/n) < 1/n$ for all $t\in K_n$ for each choice of $y\in U_{K_n}$, and this numerical integration approximates the value of $T(y)(t)$ with arbitrarily high accuracy. 

We indicate our numerical integration scheme using the formula
$$
T(y)(t) = \int_0^1G(\mathbf y(t), t, \mathbf y(s), s)ds \approx \sum_{i=0}^n w_i(t)G(\mathbf y(t), t, \mathbf y(s_i), s_i),
$$
where $s_i$ indicates the $i^{\rm th}$ grid point of $\{t_i\}\subset [0,1]$. We can therefore obtain the evaluation of $T(y)$ at the grid points $t_j$ as
$$
T(y)(t_j) = \int_0^1G(\mathbf y(t_j), t_j, \mathbf y(s), s)ds \approx \sum_{i=0}^n w_i(t_j)G(\mathbf y(t_j), t_j, \mathbf y(s_i), s_i),
$$
by choosing $t$ to be one of the grid points. 

From our regularity assumptions on the derivatives we can uniformly bound the error on evaluating $T(y)$ at the points $t_j$, so that for sufficiently dense grids the evaluation error is smaller than $\epsilon/2$ for any choice of $\epsilon>0$, when evaluating on functions $y$ in a neighborhood of $y^*$. 

Let us consider now a permutation of the input of $T(y)$ for some $\sigma \in \Sigma_n$. This means that we permute the grid points $\{t_i\}$ as $\{t_{\sigma i}\}$. The approximated integration above gives
$$
T(y)(t_i) \approx \sum_i w_{\sigma i}(t_{\sigma j})G(\mathbf y(t_{\sigma j}), t_{\sigma j}, \mathbf y(s_{\sigma i}), s_{\sigma i}) = \sum_i w_i(t_{\sigma j})G(\mathbf y(t_{\sigma j}), t_{\sigma j}, \mathbf y(s_i), s_i), 
$$
where the second equality follows from the fact that we are summing over all the permuted indices $i$. This means that our approximation of the integration, evaluated on grid points, is permutation equivariant. Applying Theorem~2 in \cite{yun2019transformers} we are able to find a transformer architecture and a weight configuration, which we denote by $\mathcal T$, such that $\|\sum_{i=0}^n w_i(t_j)G(\mathbf y(t_j), t_j, \mathbf y(s_i), s_i)-\mathcal T(y(t_j))\|_p < \epsilon/2$, as a function of the $t_j$'s. As a consequence, we obtain the approximation
\begin{eqnarray*}
    \|T(y)(t) - \mathcal T(y(t))\|_p &\leq& \|T(y)(t_j) - \sum_{i=0}^n w_i(t_j)G(\mathbf y(t_j), t_j, \mathbf y(s_i), s_i)\|_p\\
    &+& \|\sum_{i=0}^n w_i(t_j)G(\mathbf y(t_j), t_j, \mathbf y(s_i), s_i) - \mathcal T(y(t_j))\|_p\\
    &<& \epsilon/2 + \epsilon/2 = \epsilon,
\end{eqnarray*}
for any choice of $y$ in a neighborhood of $y^*$.

\section{Detailed description of the solver and the training procedure}\label{sec:solver}

We give here a detailed account of the implementation of the models NIE and ANIE ($1D$ and $(n+1)D$ IEs, respectively). More specifically, we provide a more thorough description of Algorithm~\ref{algo:NIE} and Algorithm~\ref{algo:ANIE} for solving the IEs associated to neural networks $G$ (feed-forward) and $\mathfrak {Att}$ (transformer), and contextualize these algorithms in the optimization procedure that learns the neural networks.  

\subsection{Implementation of NIE}\label{sec:NIE_solving_training}

We only consider the case of Equation~\ref{eqn:gen_NIE}, as the case where the function $G$ splits in the product of a kernel $K$ and the (possibly) nonlinear function $F$ is substantially identical. We observe that the main components of the training of NIEs are two. An optimization step that targets $G$, and a solver procedure to obtain a solution associated to the integral equation individuated by $G$, or more precisely the integral operator that $G$ defines. Therefore, we want to solve Equation~\ref{eqn:gen_NIE} for a fixed neural network $G$, determine how far this solution is from fitting the data, and optimize $G$ is such a way that at the next step we obtain a solution that more accurately fits the data. At the end of the training, we have a neural network $G$ that defines an integral operator, and in turn an integral equation, whose associated solution(s) approximates the given data. 

To fix notation, let us call $X$ the dataset for training. This contains several instances of $n$ dimensional curves with respect to time. In other words, we consider $X = \{X_i\}_{i\leq N}$ where $N$ is the number of instances and each $X_i = \{\mathbf x^i_0, \ldots, \mathbf x^i_m\}$, where each $\mathbf x_i\in \mathbb R^q$ is a $q$-dimensional vector, and the sequence of $\mathbf x^i_k$ refers to a discretization of the time interval where the curves are defined. For simplicity, we assume that time points are take in $[0,1]$. The neural network $G$ defining the integral operator will be denoted $G_{\theta}$, in order to explicitly indicate the dependence of $G$ on its parameters. The objective of the training is to optimize $\theta$ in such a way that the corresponding $G_{\theta}$ defines an IE whose solutions $\mathbf y_i(t)$ corresponding to different initializations pass through the discretized curves $\mathbf x^i(t)$. 

Let us now consider one training step $n$, where the neural network $G_{\theta_n}$ has weights obtained from the previous training steps (or randomly initialized if this is the first step). We need to solve the IE
\begin{equation}\label{eqn:IE_n}
\mathbf y = f(t) + \int_{\alpha(t)}^{\beta(t)} G_{\theta_n}(\mathbf y,t,s)ds,
\end{equation}
associated to the integral operator $\int_{\alpha(t)}^{\beta(t)} G_{\theta_n}(\mathbf y,t,s)ds$ corresponding to the weights $\theta_n$ at training step $n$.

For simplicity we consider a batch size of $1$, so that our training curve is given by $\{\mathbf x_0, \ldots, \mathbf x_m\}$, where we suppress the superscript $i$ because there is only one curve. Then, we select the first vector $\mathbf x_0$, and use this to initialize a full curve with repeated instances of this. In other words, we define $f(t) = x_0$ for all times $t$. We now apply the IE solver procedure, and set the zero order solution to the IE to be $\mathbf y^0(t) = f(t) = \mathbf x_0$. We now apply the integral operator determined by $G_\theta$ computing
$$
\mathbf z^0 = f(t) + \int_{\alpha(t)}^{\beta(t)} G_{\theta_n}(\mathbf y^0,t,s)ds.
$$
Observe, that at this stage, we can perform the integration over the interval $[\alpha(t),\beta(t)]$ for each time $t$, since $\mathbf y^0$ is given for all times $t$. We then set $\mathbf y^1 = r\mathbf y^0 + (1-r)\mathbf z^1$ where $r$ is a smoothing factor $0\leq r < 1$ which is set beforehand. The function $\mathbf y^1(t)$ is now the new approximation for the solution of the IE given in Equation~\ref{eqn:IE_n}. We can now compute the global error between $\mathbf y^0$ and $\mathbf y^1$, which we denote by $m(\mathbf y^0,\mathbf y^1)$. This error is internal to the solver and does not refer to how well the model fits the data. It refers to how far the solver is from converging. We iterate this procedure. Let us assume that this has been done $k$ times. Then, we have a function that approximates a solution of the IE at the $k^{\rm th}$ iteration, denoted by $\mathbf y^k(t)$. We compute
$$
\mathbf z^k = f(t) + \int_{\alpha(t)}^{\beta(t)} G_{\theta_n}(\mathbf y^k,t,s)ds,
$$
where, as before, we can evaluate the integral over the intervals $[\alpha(t),\beta(t)]$ because the function $\mathbf y^k(t)$ is defined over the full time length of the dynamics. 

This iterative procedure converges to a solution of the integral equation for the integral operator defined through $G_{\theta_n}$, \cite{delves1988computational}. In order to optimize the parameters $\theta$ of $G$ we require gradients on the input of $G_{\theta_n}$ when applying the neural network, we compute the loss between the solution $\mathbf y$ obtained through the iterative solution and the data, and we then backpropagate. 

\subsection{Implementation of ANIE}\label{sec:ANIE_solving_training}

We now consider ANIE, which is an IE model where the integral is approximated via self-attention. As the iterative solver procedure to obtain a solution of the IE determined by the integral operator is conceptually the same as in the case of NIE given above, we focus mostly on the details relative to the use of self-attention in this setting. Firstly, we consider an IE with space and time, which takes the form of Equation~\ref{eqn:PIE_appendix}. Our dataset now consists of instances of a given dynamics $X = \{X_i\}_{i\leq N}$, where $N$ is the number of instances in the dataset, and each $X_i = \{\mathbf x^i_{s_1,\ldots,s_d,j}\}$ is a family of $q$-dimensional vectors (where $q$ is the dimension of the dynamics), indexed by the spatial and temporal indices $s_1, \ldots, s_d$ and $j$ corresponding to a discretization (e.g. a mesh) of the spatio-temporal domain $\Omega\times [0,T]$. Observe that the dimension of the spatial domain $\Omega$ here is assumed to be $d$, therefore implying that each $\mathbf x$ depends on $d$ indices. Therefore, one can think of each dynamics instance in the dataset as being a temporal sequence of spatial meshes, e.g. a sequence of images when $d=2$. We will assume that the number of time points in such a sequence is equal to $m_T$, that the total number of space points is equal to $m_\Omega$, and we set $m= m_tm_\Omega$.

For the sake of simplicity we assume that the attention model approximating the integral operator consists of a single self-attention layer. Let $\mathfrak{Att}$ denote a self-attention layer, and assume that $\mathfrak{Att}: \mathbb R^{m\times (q+d+1)}\longrightarrow \mathbb R^{m\times (q+d+1)}$. Observe that the attention layer maps sequences of length $m$ of $(q+d+1)$-dimensional vectors to sequences of the same type. We therefore think of $\mathfrak{Att} : \mathcal X \longrightarrow \mathcal Y$ as a mapping between two function spaces $\mathcal X$ and $\mathcal Y$, whose elements are functions $\mathbf y(\mathbf x,t)$ in a discretized form, with $\mathbf x\in \Omega$ and $t\in [a,b]$. As discussed in \cite{cao2021choose} (see also \cite{xiong2021nystromformer}), the self-attention mechanism can be thought of as an approximation of an integral operator where given a discretized function $\mathbf y(\mathbf x,t)$, $\mathfrak{Att}(\mathbf y(\mathbf x,t))$ is another discretized function obtained through an approximation of an integration over the variables $\mathbf x$ and $t$. This theoretical motivation, and the computational complexity of performing Monte Carlo integration in higher dimensions, led us to consider an IE solver where instead of learning a simple neural network $G$ as in the setting of NIE, we learn the integral operator in the form of its attentional approximation $\mathfrak{Att}$. 

\begin{figure*}[ht]
\centering
  \includegraphics[width=0.6\textwidth]{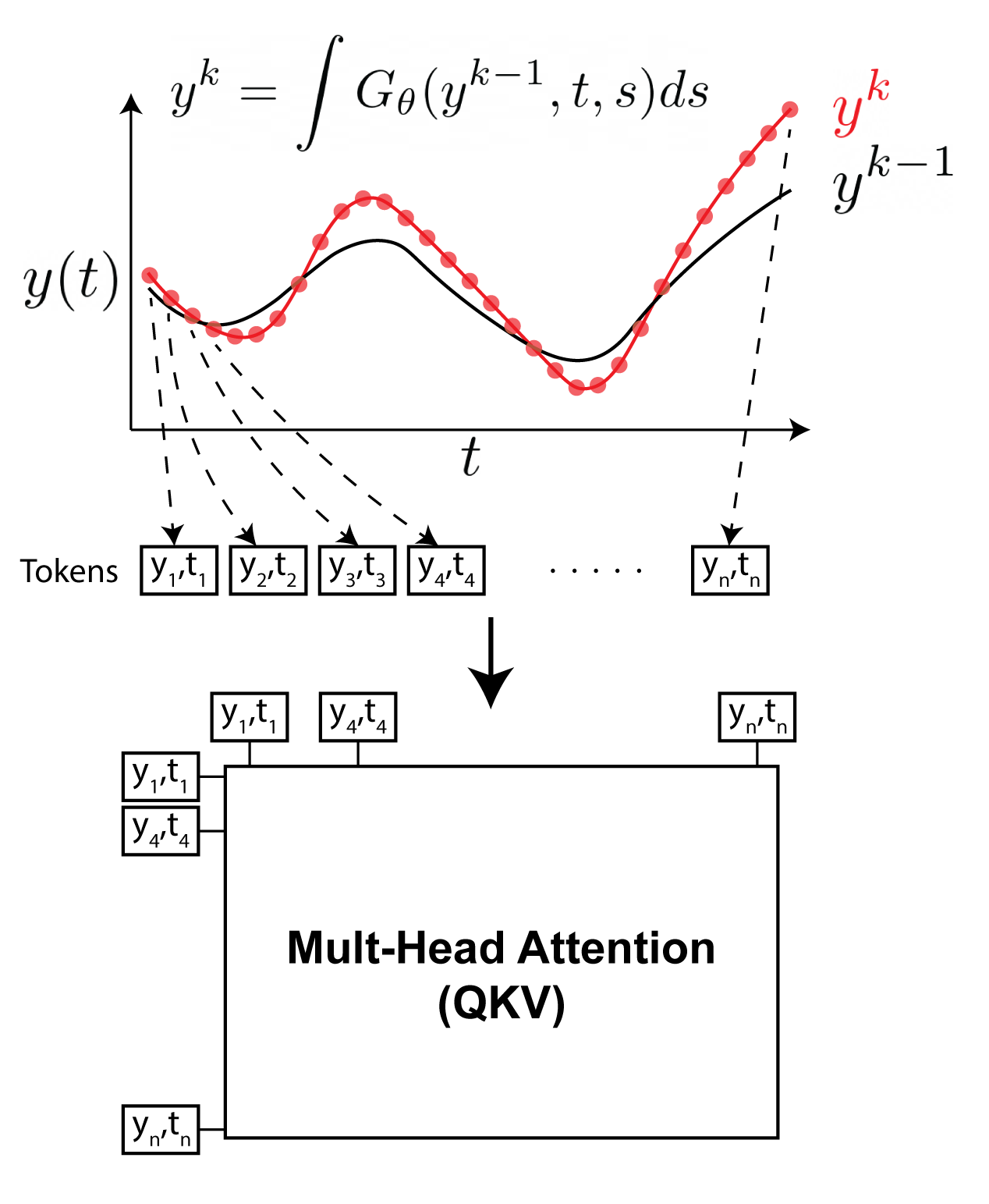}
  \caption{Integration through self-attention}\label{fig:Transformer_integration}
\end{figure*}

As for the detailed description of NIE given above, we assume that the batch size is equal to $1$, and the dataset is $X = \{X_i\}_{i\leq N}$ with $X_i = \{\mathbf x^i_{s_1,\ldots,s_d,j}\}$ for a discretization of a spatio-temporal domain $\Omega\times [0,T]$ as described at the beginning of this subsection. Let $\mathfrak {Att}_\theta$ denote the transformer with parameters $\theta$ obtained at epoch $n$ of the training session. Here if $n=0$ it simply means that $\mathfrak {Att}_\theta$ is randomly initialized. We want to inspect epoch $n+1$. The IE we solve at each training epoch takes the form
\begin{equation}\label{eqn:IE_att_n}
  \mathbf y = f(\mathbf x, t) + \mathfrak {Att}_\theta(\mathbf y, \mathbf x, t),
\end{equation}
where $\mathfrak {Att}_\theta(\mathbf y, \mathbf x, t)$ is an approximation of an integral operator $\int_{\Omega}\int_0^TG(\mathbf y,\mathbf x, \mathbf x', t, s) d\mathbf x' ds$ for some $G$. The solver is initialized through the free function $f(\mathbf x,t)$ which plays the role of first ``guess'' for the solution of the IE. Observe that since $f$ is evaluated on the full discretization of $\Omega\times [0,T]$, then the length $m$ of the sequence of vectors that approximates $f(\mathbf x,t)$ equates the product of number of space points $s_k$ for each dimensions and the time points $t_r$. 
The solver therefore creates its first approximation by setting $\mathbf y^{0}(\mathbf x,t) = f(\mathbf x,t)$. Then, the for the first iteration of the solver we create the new sequence $\tilde{\mathbf y}^0$ by concatenating to each $\mathbf y$ and the spatiotemporal $m$ coordinates $(\mathbf x_s,t_r)$. Now, $\tilde{\mathbf y}$ consists of a sequence of $m = m_Tm_\Omega$ vectors (one per spacetime point) which also are possess a spacetime encoding (through concatenation). See Figure~\ref{fig:Transformer_integration} for a schematic representation of the integration procedure through a transformer. Then, we set
$$\tilde {\mathbf y}^1(\mathbf x,t) = f(\mathbf x,t) + \mathfrak {Att}_\theta (\tilde{\mathbf y}^0),$$
where the dependence of $\tilde {\mathbf y}^1$ on spacetime coordinates $\mathbf x$ and $t$ indicates that we have one vector $\tilde y^1$ per spacetime coordinate. If $q$ is the dimension of the dynamics (i.e. the number of channels per spacetime point), then the sequence $\tilde {\mathbf y}^1$ consists of vectors of dimension $q+d+1$, where $d$ is the number of space dimensions. This happens because $\tilde {\mathbf y}^1$ is the output of a transformer of a sequence obtained by a sequence of $q$-dimensional vectors concatenated with a $(d+1)$-dimensional sequence. The two simplest options are either to discard the last $d+1$ dimensions of the vectors, or add an additional linear layer that projects from $q+d+1$ dimensions to $q$. As tests have not shown a significant difference between the two approaches, we have adopted the former. So proceeding we obtain the $1$-dimensional sequence $\mathbf Z^1(\mathbf x,t)$. Lastly, we set $\mathbf y^1(\mathbf x,t) = r\mathbf y^0 + (1-r)\mathbf z^1$, where $r$ is a smoothing factor that is a hyperparameter of the solver. One therefore computes the error $m(\mathbf y^0,\mathbf y^1)$ between the initial step and the second one to quantify the degree of change of the new approximation, where $m(\bullet,\bullet)$ is a global error metric fixed throughout.

\begin{figure*}[ht]
\centering
  \includegraphics[width=0.8\textwidth]{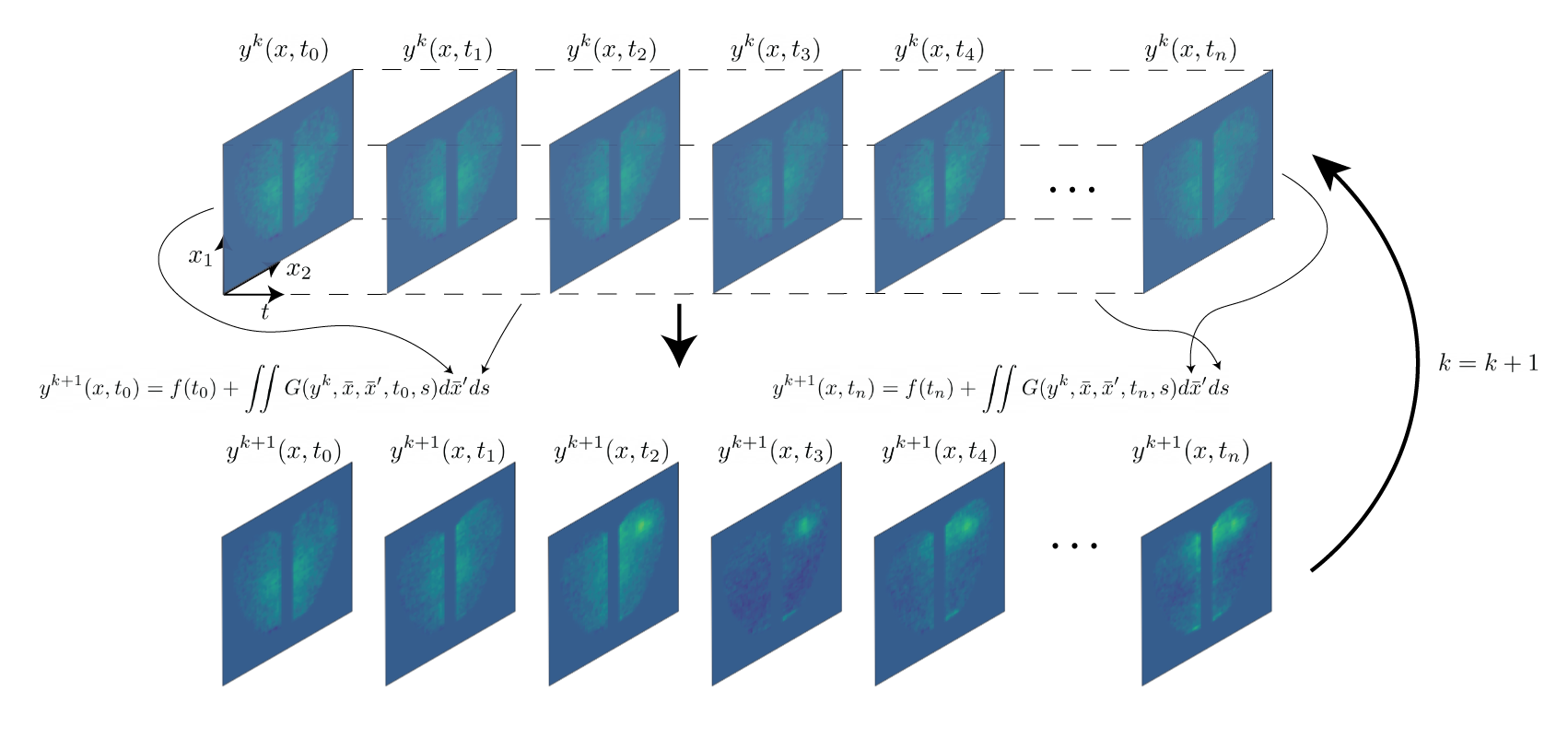}
  \caption{Solver between steps $k$ and $k+1$. The whole dynamics consists of frames ordered consecutively with respect to time. Integration with respect to space and time is performed on the $\mathbf y^k$ function to produce the guess $k+1$. Then the procedure is repeated until convergence.}\label{fig:spatial_IE}
\end{figure*}

Now we iterate the same procedure and assuming that the approximation $\mathbf y^i$ to the equation has been obtained, then we concatenate spacetime coordinates to obtain $\tilde {\mathbf y}^i$ and set
$$\tilde {\mathbf y}^{i+1}(\mathbf x,t) = f(\mathbf x,t) + \mathfrak {Att}_\theta (\tilde{\mathbf y}^i),$$
which we use to obtain $\mathbf z^{i+1}$ (by deleting the last $d+1$ dimensions). Then we set $\mathbf y^{i+1} = r\mathbf y^i + (1-r)\mathbf z^{i+1}$ and compute the global error $m(\mathbf y^i,\mathbf y^{i+1})$. Figure~\ref{fig:spatial_IE} shows a solver step integration in detail.

In practice, the number of iterations for the solver is a fixed hyperparameter which we have set between $3$ and $5$ in our experiments. This has been suffcient to achieve good results, and to learn a model that is stable under the solving procedure described above. Since the solver is fully implemented in pytorch and the model that approximates the integral operator is a transformer, we can simply backpropagate through the solver at each epoch, after we have solved for $\mathbf y$ and compared the solution to the given data $\{X_i\}_{i\leq N}$.

We complete this subsection with a more concrete description of the motivations for approximating integration through the mechanism of self-attention. Very similar perspectives have appeared in other sources (see e.g. \cite{cao2021choose,xiong2021nystromformer}), but we provide a formulation of such considerations that more easily fit the perspectives of integral operators for integral equations used in this article. This also serves as a more explicit description of the $\frak{Att}$ found in Algorithm~\ref{algo:ANIE}. 

We consider an $n$-dimensional dynamics $\mathbf y(\mathbf x, t)$ depending on space $\mathbf x \in \Omega$ (for some domain $\Omega$) and time $t \in [0,1]$. The queries, keys and values of self-attention can be considered as maps $\psi_W: \mathbb R^{n+1}\times \Omega \times [0,1] \longrightarrow \mathbb R^{1\times d}$, where $d$ is the latent dimension of the self-attention, and $W = Q, K, V$ for queries, keys and values, respectively. Then, (for $W = Q, K, V$) we have 
$$W[\mathbf y | \mathbf x | t] = \begin{bmatrix}
           \psi_W^0[\mathbf y | \mathbf x | t] \\
           \vdots \\
           \psi_W^i[\mathbf y | \mathbf x | t]\\
           \vdots \\
           \psi_W^{d-1}[\mathbf y | \mathbf x | t]
         \end{bmatrix},$$
where $[\mathbf y | \mathbf x | t]$ indicates concatenation of the terms in the bracket. Let us consider now the “traditional” quadratic self-attention. Similar considerations also apply for the linear attention used in the experiments, mutatis mutandis. The product between queries and keys gives 
$$[( \cdots \psi_Q^i[\mathbf y | \mathbf x | t] \cdots)\cdot ( \cdots \psi_K^j[\mathbf y | \mathbf x | t] \cdots)^T]_{ij} = (\psi_Q^i\cdot \hat \psi_K^j),$$ where $T$ indicates transposition, and $\hat \psi$ indicates the columns of the transposed matrix. Then, if $z$ is the output of the self-attention layer (observe that this consists of $(\mathbf z_i)_j$ where $i$ indicates a spatiotemporal point, and $j$ indicates the $j^{th}$-dimension of the $n$-dimensional dynamics. Then, we have 
$$(\mathbf z_i)_j = \sum_j (\psi_Q^i\cdot \hat \psi_K^j)\psi_V^j \approx \int_{\Omega\times [0,1]} K(\mathbf y, \mathbf x, t, \mathbf y', \mathbf x', t')F(\mathbf y', \mathbf x', t') d\mathbf x' dt',$$ 
where the prime symbols indicates the variables we are summing upon (this is why the are “being integrated” in the integral), and $\mathbf y$ is evaluated at $\mathbf x, t$, while $\mathbf y'$ is evaluated at $\mathbf x', t'$.

\subsection{Dependence of the model on iteration steps}

\begin{figure*}[ht]
\centering
  \includegraphics[width=0.8\textwidth]{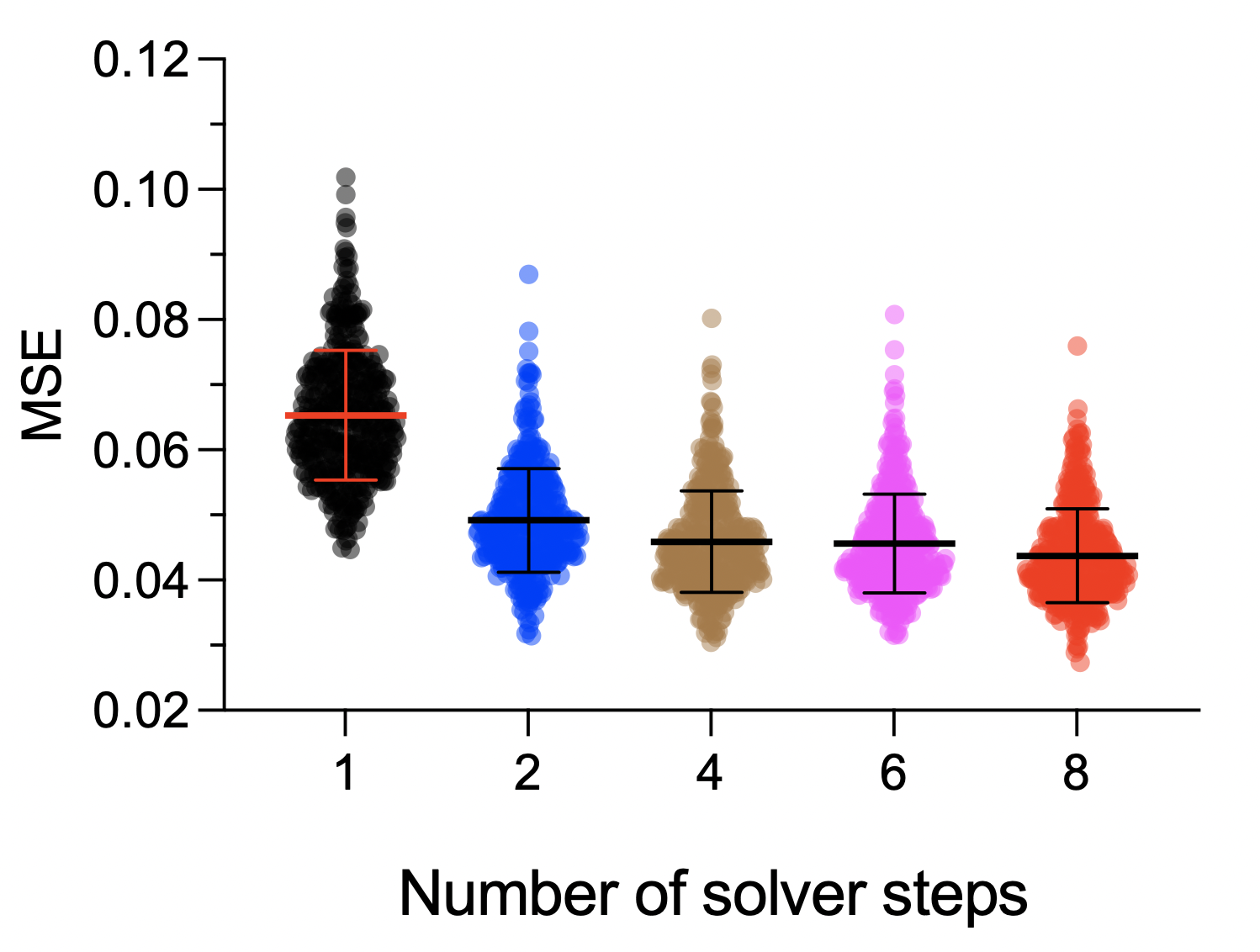}
  \caption{Dependence of the model's fit with respect to number of iterations for a single layer ANIE architecture. The figure represents the results in Table~\ref{tab:iterations_1}.}\label{fig:iterations}
\end{figure*}

We explore here the dependence of model extrapolation on initial condition for the Navier-Stokes dataset with respect to the number of iterations of the solver. The results are reported in Table~\ref{tab:iterations_1} and Table~\ref{tab:iterations_4}, where mean squared error and standard deviations are reported. Figure~\ref{fig:iterations} shows the results reported in Table~\ref{tab:iterations_1}. We perform our experiments with two different models, one with a much higher number of parameters than the other. We see that for a smaller model, the impact of the number of solver steps becomes much more pronounced. This indicates that while a very large model is able to compensate the effect of the solver steps and reduce the difference in testing quality, a smaller model can greatly benefit from a higher number of iterations. We notice, in particular, that an ANIE model with a single layer performs as well as an ANIE model with $4$ layers and lower number of iterations. In all cases, higher number of solver steps give better evaluations than single iteration models with statistical significance ($P<0.0001$). 

\begin{table*}
\caption{Dependence of the model's fit with respect to change in number of iterations of the solver for a single layer architecture of ANIE. Mean squared error and standard deviations are reported. For higher iterations the error during testing decreases in a statistically significant manner ($P<0.0001$)}\label{tab:iterations_1}
\centering

\resizebox{\textwidth}{!}{\begin{tabular}{|c | c | c | c | c |}
  \hline 
  iter $= 1$ & iter $= 2$ & iter $= 4$ & iter $= 6$ & iter $= 8$\\
  \hline 
    $0.06532\pm 0.00996$ & $0.04918\pm 0.00794$ & $0.04591\pm 0.00780$ & $0.04563\pm 0.00760$ & $0.04371\pm 0.00719$\\
  \hline
\end{tabular}}

\end{table*}

\begin{table*}
\caption{Dependence of the model's fit with respect to change in number of iterations of the solver for a $4$ layers architecture of ANIE. Mean squared error and standard deviations are reported. For higher iterations the error during testing decreases in a statistically significant manner ($P<0.0001$)}\label{tab:iterations_4}
\centering

\resizebox{\textwidth}{!}{\begin{tabular}{|c | c | c | c | c |}
  \hline 
  iter $= 1$ & iter $= 2$ & iter $= 3$ & iter $= 4$ & iter $= 5$\\
  \hline 
    $0.04485\pm 0.00766$ & $0.04417\pm 0.00764$ & $0.04359\pm 0.00773$ & $0.04229\pm 0.00711$ & $0.04196\pm 0.00714$\\
  \hline
\end{tabular}}

\end{table*}

\section{Computational cost}

We now give more details regarding the computational cost of our models. 

The theoretical order of the computation for NIE per iteration is in the order of $N\times T$, where $N$ is the number of Monte Carlo sampling points, and $T$ is the number of time points used in the solver. This has to be multiplied by the number of iterations, which for example has been taken to be $3$ in the experiments on training speed.

For ANIE we have performed our experiments using a linear version of the self-attention, which requires a linear computational cost in the number of spacetime points used (this changes depending on the resolution of the dataset). So, for a spacetime grid $\Omega_n \subset \Omega$ consiting of $n$ space points, and a grid $T_m \subset I$ consisting of $m$ time points, the computational cost is in the order of $n\cdot m$ times the number of solver itrations. The iterations for ANIE varied between $3$ and $7$ throuoghout the experiments. We observe that quadratic attention would result in a computational cost of the order of $(nm)^2\cdot r$, where $r$ is the number of iterations of the solver.

\section{Artificial Dataset Generation}\label{data_generation}

\subsection{Lotka-Volterra System}

\begin{figure*}[ht]
\centering
  \includegraphics[width=0.8\textwidth]{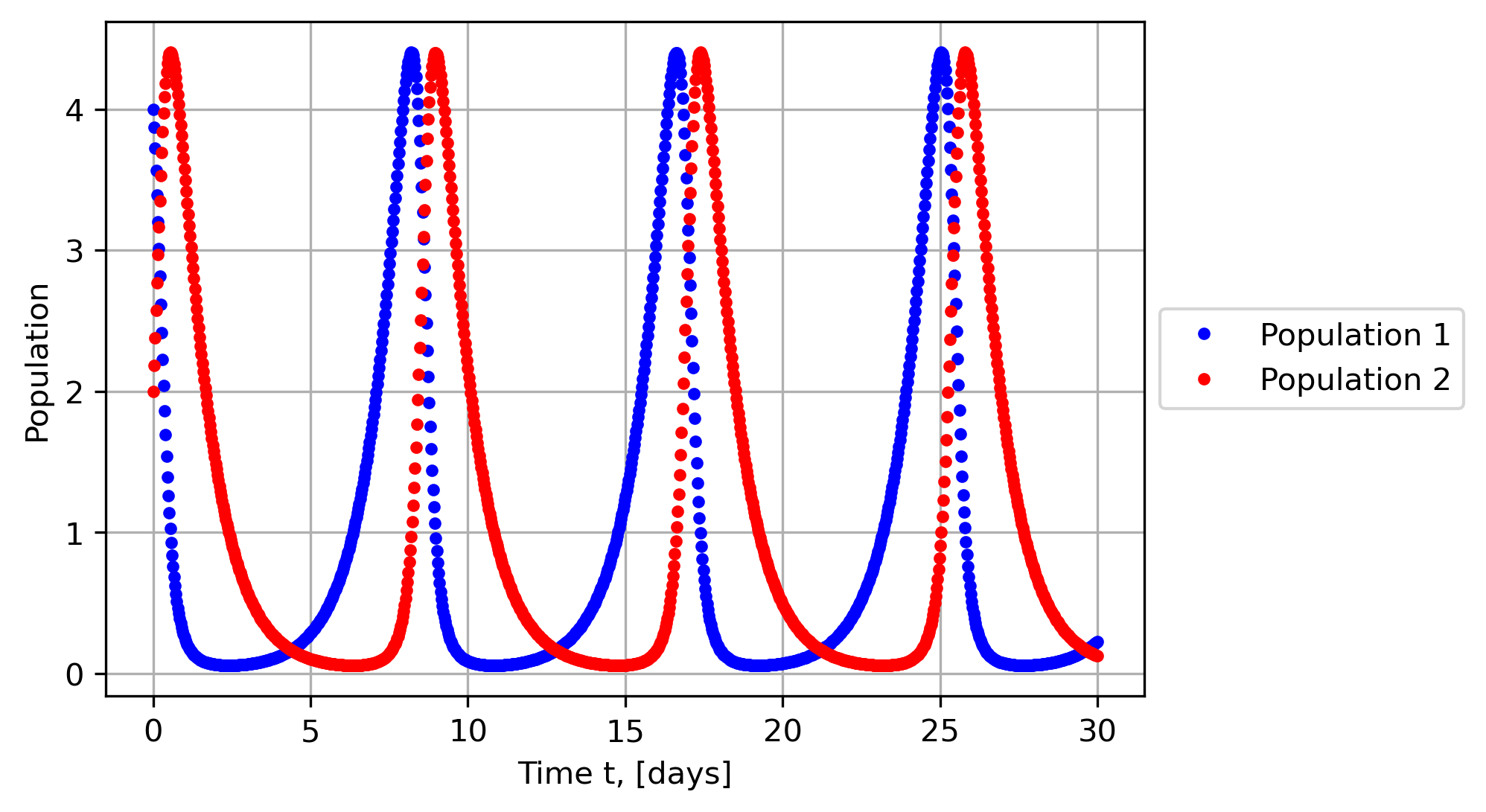}\caption{Example visualization of Lotka-Volterra 2D System}\label{fig:Lotka-Volterra}
\end{figure*}

The Lotka-Volterra equations are a classic system of nonlinear differential equations that model the interaction between two populations. The equations are given by:
\begin{equation*}
\centering
\frac{dx}{dt} = \alpha x y - \beta y
\end{equation*}
\begin{equation*}
\centering
\frac{dy}{dt} = \delta x y - \gamma y
\end{equation*}
where $\alpha$ and $\delta$ define the population interaction terms, and $\beta$ and $\gamma$ are the intrinsic population growth for population $x$ and $y$. To generate our dataset, 100 values of $\alpha, \beta, \delta$ and $\gamma$ have been randomly generated and the corresponding system has been solved with a fixed initial condition. Our code was adapted from \url{https://scipy-cookbook.readthedocs.io/items/LoktaVolterraTutorial.html}. An example visualization of a solution is given in Figure~\ref{fig:Lotka-Volterra}.

\subsection{Lorenz System}

The Lorenz system is a 3-dimensional system of ordinary differential equations, for modelling atmosferic convection. Furthermore, this system is known to be chaotic, which means that small variations of initial conditions can significantly affect the final trajectory. The system is given by: 

\begin{equation*}
\centering
\frac{dx}{dt} = \sigma (y - x)
\end{equation*}
\begin{equation*}
\centering
\frac{dy}{dt} = x (\rho - z) - y
\end{equation*}
\begin{equation*}
\centering
\frac{dz}{dt} = xy - \beta z
\end{equation*}

\begin{figure*}[ht]
\centering
  \includegraphics[width=0.7\textwidth]{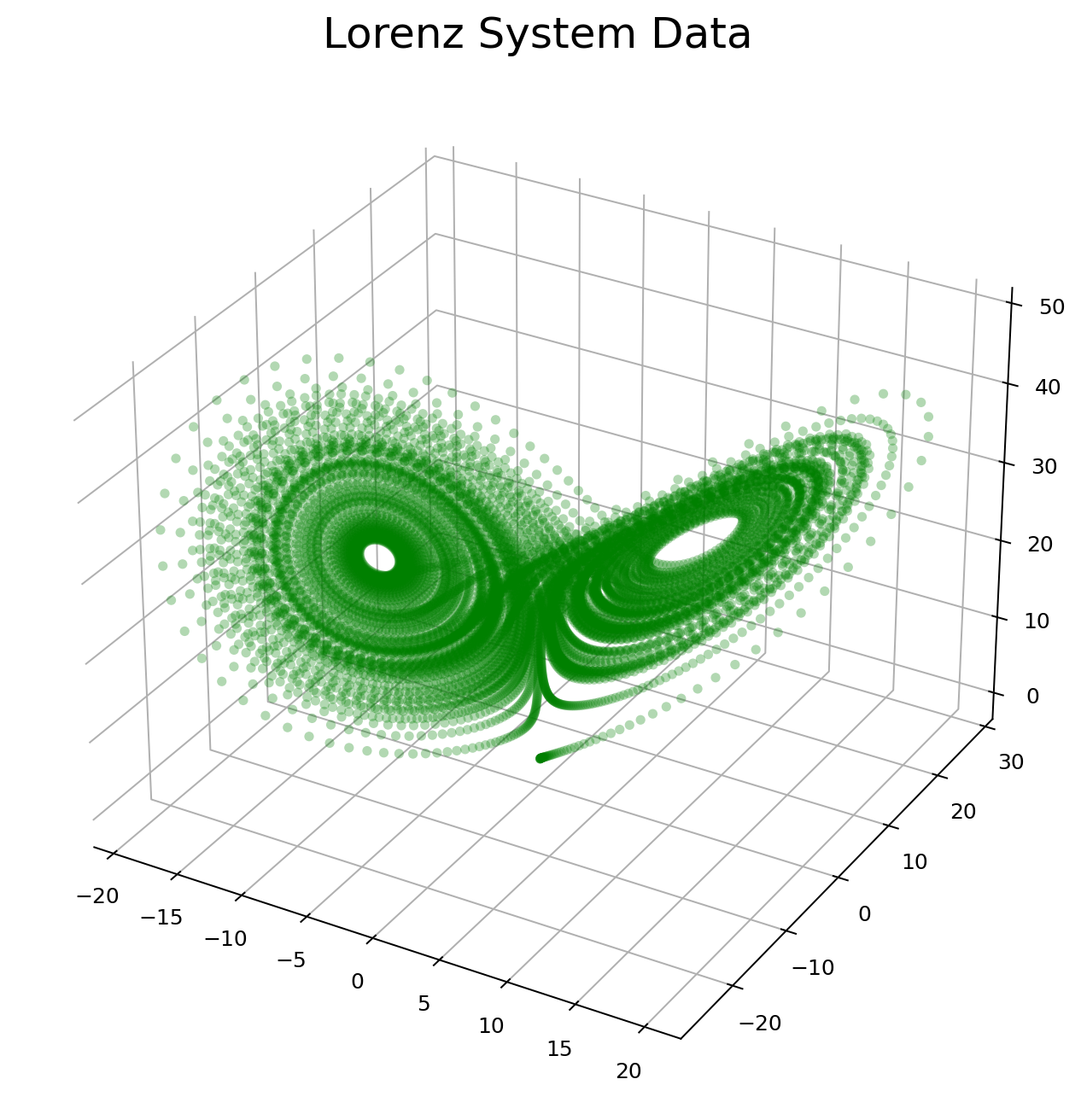}\caption{Example visualization of 3D Lorenz Dynamical System}\label{fig:3d_lorenz}
\end{figure*}

We have sampled 100 random initial conditions, and have solved the system with the same fixed parameters. Our code was adapted from \url{https://github.com/gboeing/lorenz-system}. An example visualization with default hyperparams is given in Figure~\ref{fig:3d_lorenz}.

\subsection{Integral Equation Spirals}\label{sec:IE_spirals}

The $2D$ integral equation spirals have been obtained by solving an IE with the following form:
\begin{eqnarray*}
       \mathbf y(t) = \int_0^t \begin{bmatrix}
           \cos 2\pi(t-s) & -\sin 2\pi(t-s) \\
           -\sin 2\pi(t-s) & -\cos 2\pi(t-s)
       \end{bmatrix} \tanh(2\pi \mathbf y(s))dx
       + \mathbf z_0 + \begin{bmatrix}
                \cos(t) \\
                \cos(t+\pi)
       \end{bmatrix},   
\end{eqnarray*}
where $z_0$ was sampled from a uniform distribution.

\begin{figure*}[ht]
\centering
  \includegraphics[width=0.5\textwidth]{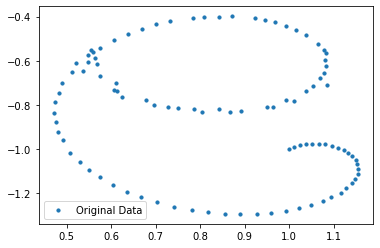}\caption{Example visualization of 2D Integral Equation Spiral}\label{fig:2d_spiral}
\end{figure*}

The equation has been solved numerically through our solver (with analytical functions instead of neural networks) for different known functions $f$ corresponding to different choices of $\mathbf z_0$. An example visualization of a solution is given in Figure~\ref{fig:2d_spiral}.  

\subsection{FMRI data generation}\label{sec:fMRI_data}
The simulated fMRI data was generated using \textit{neurolib} \cite{cakan2021neurolib}. The authors of this tool provide code to generate fMRI data for Resting-state with a given structural connectivity matrix and a delay matrix. The code can be found in their GitHub page \footnote{https://github.com/neurolib-dev/neurolib/blob/master/examples/example-0-aln-minimal.ipynb}. We used this code to generate 100000 time points of data for 80 voxels corresponding to regions of the cortex. 

The generated data is normalized via computing the z-score of the logarithm of the whole data. This data is then downsampled in time by a factor of 10, thus resulting in 10k time points. In our tests, we use the first 5k points, where the first 2.5k points are used for training and the remaining points are reserved for testing. During batching, each point is taken as the initial condition of a curve of length 20 points. 

\section{Calcium imaging dataset}
\label{appendix:calcium_imaging}
C57BL/6J mice were kept on a 12h light/dark cycle, provided with food and water ad libitum, and housed individually following headpost implants. Imaging experiments were performed during the light phase of the cycle. For mesoscopic imaging, brain-wide expression of jRCaMP1b was achieved via postnatal sinus injection as described in  \cite{barson2020simultaneous,hamodi2020transverse}. 

Briefly, P0-P1 litters were removed from their home cage and placed on a heating pad. Pups were kept on ice for 5 min to induce anesthesia via hypothermia and then maintained on a metal plate surrounded by ice for the duration of the injection. Pups were injected bilaterally with 4 ul of AAV9-hsyn-NES-jRCaMP1b ($2.5\times10^{13}$ gc/ml, Addgene). Mice also received an injection of AAV9-hsyn-ACh3.0 to express the genetically encoded cholinergic sensor ACh3.0, (Jing et al., 2020, although these data were not used in the present study. Once the entire litter was injected, pups were returned to their home cage.

Surgical procedures were performed on sinus-injected animals once they reached adulthood ($>$P50). Mice were anesthetized using 1-2\% isoflurane and maintained at 37ºC for the duration of the surgery. For mesoscopic imaging, the skin and fascia above the skull were removed from the nasal bone to the posterior of the intraparietal bone and laterally between the temporal muscles. The surface of the skull was thoroughly cleaned with saline and the edges of the incision secured to the skull with Vetbond. A custom titanium headpost for head fixation was secured to the posterior of the nasal bone with transparent dental cement (Metabond, Parkell), and a thin layer of dental cement was applied to the entire dorsal surface of the skull. Next, a layer of cyanoacrylate (Maxi-Cure, Bob Smith Industries) was used to cover the skull and left to cure ~30 min at room temperature to provide a smooth surface for trans-cranial imaging.

Mesoscopic calcium imaging was performed using a Zeiss Axiozoom with a 1x, 0.25 NA objective with a 56 mm working distance (Zeiss). Epifluorescent excitation was provided by an LED bank (Spectra X Light Engine, Lumencor) using two output wavelengths: 395/25 (isosbestic for ACh3.0, Lohani et al., 2020) and 575/25nm (jRCaMP1b). Emitted light passed through a dual camera image splitter (TwinCam, Cairn Research) then through either a 525/50 (ACh3.0) or 630/75 (jRCaMP1b) emission filter (Chroma) before it reached two sCMOS cameras (Orca-Flash V3, Hamamatsu). Images were acquired at 512x512 resolution after 4x pixel binning. Each channel was acquired at 10 Hz with 20 ms exposure using HCImage software (Hamamatsu).

For visual stimulation, sinusoidal drifting gratings (2 Hz, 0.04 cycles/degree were generated using custom-written functions based on Psychtoolbox in Matlab and presented on an LCD monitor at a distance of 20 cm from the right eye. Stimuli were presented for 2 seconds with a 5 second inter-stimulus interval

Imaging frames were grouped by excitation wavelength (395nm, 470nm, and 575nm) and downsampled from 512$\times$512 to 256$\times$256 pixels. Detrending was applied using a low pass filter (N=100, $f_{cutoff}=$0.001Hz). Time traces were obtained using $(\Delta F/F)_i=(F_i-F_{(i,o)} )/F_{(i,o)}$ where $F_i$ is the fluorescence of pixel $i$ and $F_{(i,o)}$ is the corresponding low-pass filtered signal.

Hemodynamic artifacts were removed using a linear regression accounting for spatiotemporal dependencies between neighboring pixels.  We used the isosbestic excitation of ACh3.0 (395 nm) co-expressed in these mice as a means of measuring activity-independent fluctuations in fluorescence associated with hemodynamic signals.  Briefly, given two $p\times1$  random signals $y_1$ and $y_2$ corresponding to $\Delta F/F$ of $p$ pixels for two excitation wavelengths “green” and "UV", we consider the following linear model:

\begin{eqnarray}
	y_1=x+z+\eta,
\end{eqnarray}
\begin{eqnarray}
	y_2=Az+\xi,
\end{eqnarray}	

where x and z are mutually uncorrelated $p\times1$ random signals corresponding to $p$ pixels of the neuronal and hemodynamic signals, respectively. $\eta$ and $\xi$ are white Gaussian $p\times1$ noise signals and A is an unknown $p\times p$ real invertible matrix. We estimate the neuronal signal as the optimal linear estimator for $x$ (in the sense of Minimum Mean Squared Error):

\begin{eqnarray}
	\hat{x} &=& H\left(\begin{array}{c} y_1  \\ y_2  \end{array}\right),\\
	H &=& \sum_{xy}{\sum_{y}}^{-1}
\end{eqnarray}

where $y=\begin{pmatrix} y_1  \\ y_2  \end{pmatrix}$ is given by stacking $y_1$  on top of $y_2$,  $\sum_y=E[yy^T ]$ is the autocorrelation matrix of $y$ and $\sum_{xy}=E[xy^T ]$ is the cross-correlation matrix between $x$ and $y$. The matrix $\sum_y$ is estimated directly from the observations, and the matrix $\sum_{xy}$ is estimated by:

\begin{eqnarray}
	\sum_{xy}=\Biggl(\sum_{y_1}-\sigma_{\eta}^{2}I- \biggl(\sum_{y_1 y_2} {\Bigl(\sum_{y_2}-\sigma_{\xi}^{2}I \Bigl)}^{-1} {\sum_{y_2}}^{-1} {\sum_{y_1 y_2}}^T \biggl)^T&0\Bigg) 
\end{eqnarray}

where $\sigma_\eta^2$ and $\sigma_\xi^2$  are the noise variances of $\eta$ and $\xi$, respectively, and $I$ is the $p\times p$ identity matrix. The noise variances $\sigma_\eta^2$ and $\sigma_\xi^2$  are evaluated according to the median of the singular values of the corresponding correlation matrices  $\sum_{y_1}$and $\sum_{y_2}$.  This analysis is usually performed in patches where the size of the patch, $p$, is determined by the amount of time samples available and estimated parameters. In the present study, we used a patch size of $p=9$.   The final activity traces were obtained by z-scoring the corrected $\Delta F/F$ signals per pixel. The dimensionality of the resulting video is then reduced via PCA to 10 components, which represents $\approx 80\%$ of data variance.

\section{Burgers' equations}\label{burgers_data}

The Burgers' equation is a quasilinear parabolic partial differential equation that takes the form 

\begin{equation}
    \frac{\partial u}{\partial t} + u\frac{\partial u}{\partial x} = \nu \frac{\partial^2 u}{\partial x^2},
\end{equation}
where $x$ is a spatial dimension, while $t$ indicates time, and $\nu$ is a diffusion coefficient called {\it viscosity} see \cite{benton1972table}. A very interesting behavior of the solutions of the Burgers' equation regards the presence of shock waves.

Our dataset is generated using the Matlab code used in \cite{li2020fourier}, which can be found in their GitHub page 
\footnote{https://github.com/zongyi-li/fourier\_neural\_operator/tree/master/data\_generation/burgers}. The solution is given on a spatial mesh of $1024$ and $400$ time points are generated from a random initial condition. We use $1000$ curves for training and test on $200$ unseen curves, where the interval spans $1/4$ of the original time used for testing.  

\section{Navier-Stokes equations}\label{navier_stokes_data}
The Navier-Stokes equations are partial differential equations that arise in fluid mechanics, where they are used to describe the motion of viscous fluids. They are derived from the conservation laws (for momentum and mass) for Newtonian fluids subject to an external force with the addition of pressure and friction forces, where the unknown function indicates the velocity vector of the fluid \cite{chorin1968numerical,fefferman2000existence}. Their expression is given by the system

\begin{equation}
    \frac{\partial}{\partial t}u_i + \sum_j u_j\frac{\partial u_i}{\partial x_j} = \nu \Delta u_i - \frac{\partial p}{\partial x_i} + f_i(\bold{x},t)
\end{equation}
\begin{equation}
    {\rm div} u = \sum_i \frac{\partial u_i}{\partial x_i}
\end{equation}
where $\Delta$ is the Laplacian operator, $f$ is the external force, and $\bold u$ is the unknown velocity function. We experiment on the same data set for $\nu = 1e-3$ of \cite{li2020fourier}, which can be found in their GitHub page \footnote{https://github.com/zongyi-li/fourier\_neural\_operator/tree/master/data\_generation/navier\_stokes}. They solved the viscous, incompressible $2D$ Navier-Stokes equation for vorticity on the unit torus, hence with periodic boundary conditions. The initial time point is sampled from a gaussian distribution. The forcing term is a linear combination of sine and cosine functions depending only on space, and independent of time. The numerical method for the solution of the equation is pseudospectral, for the vorticity-streamfunction formulation. The solver scheme follows the steps: (1) solving the Poisson equation, (2) vorticity is differentiated, (3) the non-linear term is added. A Crank-Nicholson update is used to advance time. Details can be found in Appendix A.3.3 of \cite{li2020fourier}. 

We use $4000$ instances for training and $1000$ for testing.
In our tasks, we utilize a single time point to inizialize our model (ANIE) and obtain the full dynamics from a single frame. For comparison, we use the minimal number of time points allowed for the other models for comparison. This is not always possible, for instance, FNO3D cannot be applied on a single time point, or few time points, as the time convolution needs several time points to produce significant results. Despite this significant advantage given to FNO3D, ANIE (ours) still better performs on the prediction of $10$ and $20$ time points. 

\section{Additional details on experiments and computational resources}\label{sec:exp_details}

The number of parameters for the models used in the experiments and are given in Tables \ref{tab:Hyperparameters} and \ref{tab:Hyperparameters2}. In all cases, the optimizer ``Adam'' has been employed. Experiments have been run on a 16GB NVIDIA A100 GPU. 


\begin{table*}
	\caption{Number of parameters for fMRI and 2D IE curves experiments}
	\label{tab:Hyperparameters}
	\centering
	\begin{tabular}{|l|c|c|c|c|}
		\hline
		 & ANIE & NODE & LSTM & Residual Net \\
		\hline
		Generated fMRI & $321,233$  & $319,280$ & $328,400$ & $319,280$\\
		\hline
		2D curves  &  $18,819$ & $1,654$ & $20,862$  & - \\
		\hline
	\end{tabular}
\end{table*}


\begin{table*}
	\caption{Number of parameters for PDE experiments. The ViT models and ResNet have variable number of parameters for the Burgers' equation, depending on the space and time resolutions. We report minimum and maximum number of parameters across the models.}
	\label{tab:Hyperparameters2}
	\centering
	\begin{tabular}{|l|c|c|}
            \hline 
           &  Burgers & Navier-Stokes\\
           \hline 
           LSTM & $-$ & $302,059,520$ \\
		\hline
		 FNO3D & $-$ & $6,558,537$ \\
            \hline
           Galerkin & $511,329$ &  $-$ \\
            \hline
            Conv1DLSTM & $149,520$ & $-$\\
            \hline 
            ResNet & $425,056-578,912$ & $-$ \\
            \hline
            Conv2DLSTM & $-$ & $447,528$ \\
            \hline 
            ViT & $3,823,904-6,467,872$& $105,258,016$\\
            \hline
            ViTsmall & $4,497,828-6,483,876$& $105,321,636$ \\
            \hline 
            ViTparallel & $6,979,328-9,623,296$ & $126,260,224$\\
            \hline
            ViT3D & $-$ & $105,340,096$ \\
            \hline
            ANIE & $161,154$ & $1,278,627$\\
            \hline 
	\end{tabular}
\end{table*}

\end{document}